\documentclass[journal]{IEEEtran}% Comment this line out if you need a4paper

\IEEEoverridecommandlockouts                              % This command is only needed if 
\pdfoutput=1                                                       % you want to use the \thanks command

\usepackage{pdfpages}
\usepackage{graphicx}
\usepackage{graphics} % for pdf, bitmapped graphics files
\usepackage{hyperref}
\usepackage{amsmath,amssymb}
\usepackage{bm}
\usepackage{bbold}
\usepackage{multicol}
\usepackage{glossaries}
\usepackage[]{algorithm2e}
\usepackage{subcaption}
\usepackage{optidef}
\usepackage[shortlabels]{enumitem}
\usepackage{xcolor,colortbl}
\usepackage{siunitx}

\definecolor{LightCyan}{rgb}{0.88,1,1}

\newacronym{HyQ}{HyQ}{Hydraulically actuated Quadruped}
\newacronym{DoFs}{DoFs}{Degrees of Freedom}
\newacronym{EoM}{EoM}{Equation of Motion}
\newacronym{CoM}{CoM}{Center of Mass}
\newacronym{CoMP}{CoMP}{Center of Mass Projection}
\newacronym{icp}{ICP}{Instantaneous Capture Point}
\newacronym{vrp}{VRP}{Virtual Repellent Point}
\newacronym{cop}{CoP}{Center of Pressure}
\newacronym{grf}{GRF}{Ground Reaction Force}
\newacronym{grp}{GRP}{Ground Reference Point}
\newacronym{lp}{LP}{Linear Program}
\newacronym{qp}{QP}{Quadratic Programs}
\newacronym{cwc}{CWC}{Contact Wrench Cone}
\newacronym{awp}{AWP}{Actuation Wrench Polytope}
\newacronym{zmp}{ZMP}{Zero-tilting Moment Point}
\newacronym{fwp}{FWP}{Feasible Wrench Polytope}
\newacronym{gws}{GWS}{Grasp Wrench Space}
\newacronym{wfw}{WFW}{Wrench-Feasible Workspace}
\newacronym{fsw}{FSW}{Feasible Solution of Wrench}
\newacronym{FWP}{FWP}{Feasible Wrench Polytope}
\newacronym{AWP}{AWP}{Actuation Wrench Polytope}
\newacronym{to}{TO}{Trajectory Optimization}
\newacronym{lip}{LIP}{Linear Inverted Pendulum}
\newacronym{slip}{SLIP}{Spring Loaded Inverted Pendulum}
\newacronym{cdpr}{CDPR}{Cable-Driven Parallel Robots}
\newacronym{IP}{IP}{Iterative Projection}
\newacronym{SIP}{SIP}{Sequential Iterative Projection}
\newacronym{SOCP}{SOCP}{Second Order Cone Program}

\newcommand{\eref}[1]{(\ref{#1})}

\newcommand{\eg}{\emph{e.g.,~}}
\newcommand{\etal}{\emph{et al.~}}
\newcommand{\ie}{\emph{i.e.,~}}

\newcommand{\nth}[1]{\ensuremath {#1}^\text{th}}
\newcommand{\diag}{\ensuremath \mathrm{diag}}

\newcommand{\Rnum}{\mathbb{R}} % Symbol fo the real numbers set

\newcommand{\vc}[1]{\mathbf{\bm{#1}}} 					% Vector symbol
\DeclareMathOperator*{\argmin}{\arg\!\min}				% argmin
\DeclareMathOperator*{\st}{s.t.}						% subject to
\newcommand{\mat}[1]{\ensuremath{\begin{bmatrix}#1\end{bmatrix}}}	% matrix
\ifCLASSINFOpdf
\else
\fi
\begin{document}

	\title{Feasible Region: an Actuation-Aware Extension of the Support Region}
	
	\author{Romeo Orsolino$^{1,2}$, Michele Focchi$^{1}$, St{\'e}phane Caron$^{3}$, Gennaro Raiola$^{1,4}$, Victor Barasuol$^{1}$,\\
	 Darwin G. Caldwell$^{1}$ and Claudio Semini$^{1}$% <-this % stops a space
		\thanks{This work was supported by Istituto Italiano di Tecnologia.}% <-this % stops a space
		\thanks{$^{1}$Dynamic Legged Systems (DLS) lab, Istituto Italiano di Tecnologia, Genova, Italy.
		{\tt\footnotesize email: \{michele.focchi, victor.barasuol, darwin.caldwell, claudio.semini\}@iit.it}}%
		\thanks{$^{2}$Dynamic Robot Systems (DRS), Oxford Robotics Institute (ORI), Univeristy of Oxford, UK.
		{\tt\footnotesize email: rorsolino@robots.ox.ac.uk}}%
		\thanks{$^{3}$ANYbotics AG, Hagenholzstrasse 83a, 8050, Zurich. The author was at Laboratoire  d'Informatique,  de  Robotique
			et  de  Microelectronique  de  Montpellier  (LIRMM),  CNRS-University  of
			Montpellier, Montpellier, France, at the time of contributing to this manuscript.}
		\thanks{$^{4}$Jet Propulsion Laboratory (JPL), NASA, USA
			{\tt\footnotesize email: gennaro.raiola@jpl.nasa.gov}}
		%\thanks{Scientific contribution: R.O (60\%), M.F. (30\%), S.C. (10\%), G.R. (0\%), V.B. (0\%) and C.S. (0\%). G.R., V.B. and C.S. contributed with technical support, review and funding.}
		\thanks{R.O. and M.F. developed most of the scientific contribution: including theoretical formalism, analytic calculations, numerical simulations and hardware experiments. S.C. and V.B. provided theoretical and technical support to the development of the algorithms. G.R. contributed with technical support to the software. D.G.C. and C.S. provided funding and supervised the project. R.O. took the lead in writing the manuscript. M.F., S.C., V.B. and C.S.  provided critical feedback and helped shape the manuscript.}
	}
	
	% The paper headers TODO
	%\markboth{Journal of \LaTeX\ Class Files,~Vol.~, No.~, January~2019}%

	\null
	\includepdf[pages=-]{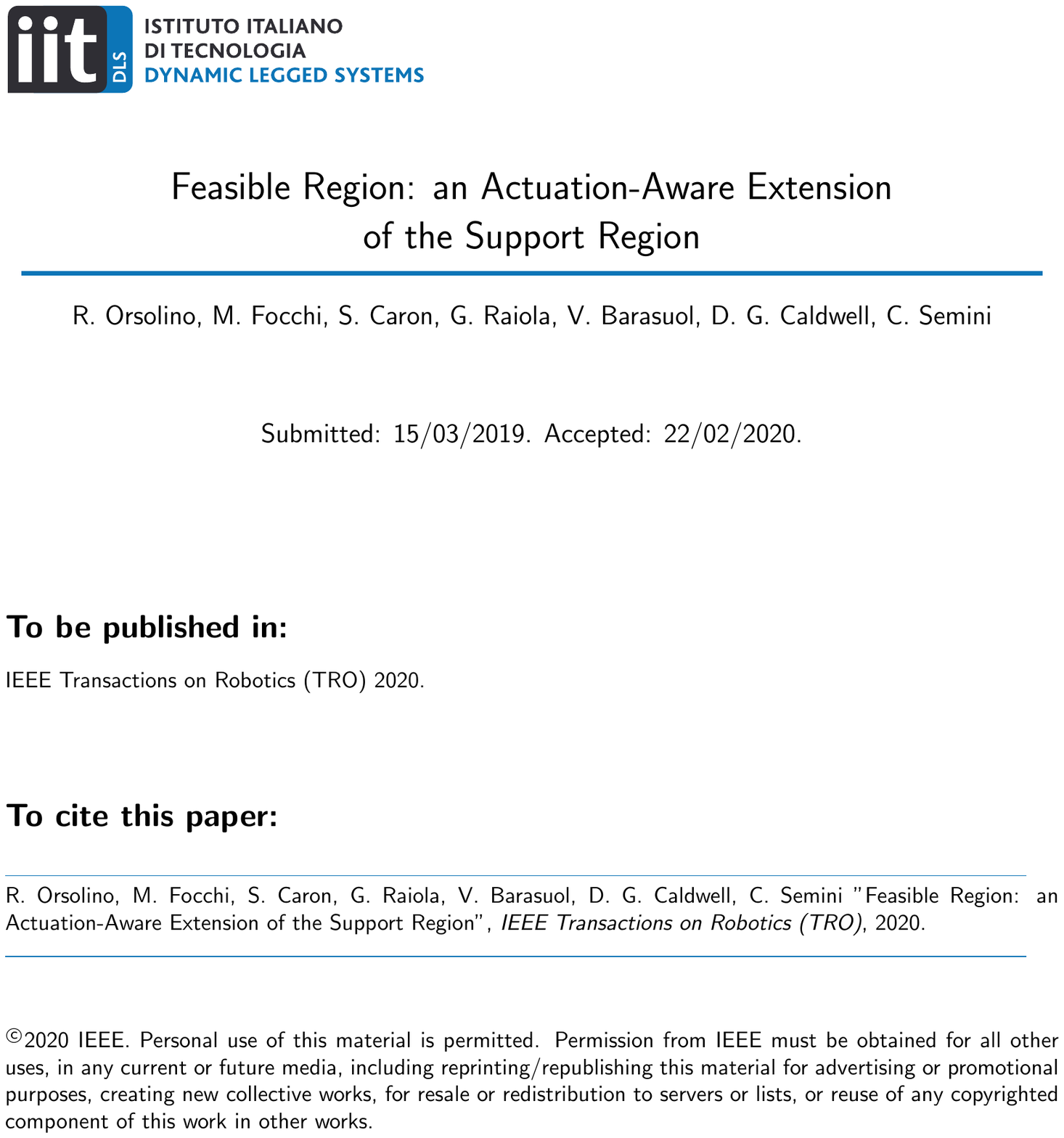}
	\maketitle
	
	\begin{abstract}
In legged locomotion the projection of the robot's \gls{CoM} being inside the convex hull of the contact points is a commonly accepted sufficient condition to achieve static balancing. However, some of these configurations cannot be realized because the joint-torques required to sustain them would be above their limits (actuation limits). In this manuscript we rule out such configurations and define the \emph{Feasible Region}, a revisited support region that guarantees both global static stability in the sense of tip-over and slippage avoidance and of existence of a set of joint-torques that are able to sustain the robot's body weight. We show that the feasible region can be employed for the online selection of feasible footholds and \gls{CoM} trajectories to achieve statically stable locomotion on rough terrains, also in presence of load-intensive tasks. Key results of our approach include the efficiency in the computation of the feasible region using an \gls{IP} algorithm and the successful execution of hardware experiments on the HyQ robot, that was able to negotiate obstacles of moderate dimensions while carrying an extra \SI[inter-unit-product =\ensuremath{\cdot}]{10}{\kilogram} payload. 
\end{abstract}
	
	\begin{IEEEkeywords}
	legged locomotion, multi-contact motion planning and control, dynamics, optimization and optimal control.
	\end{IEEEkeywords}

	% \fi
	\IEEEpeerreviewmaketitle

\section*{Supplementary Material}
{
\begin{itemize}
	\item Video of simulation and hardware results is available at: 
	\href{https://youtu.be/9pvWO2Qmo9k}{\texttt{https://youtu.be/9pvWO2Qmo9k}}
	\item Code available (python) at:\\
	\href{https://github.com/orsoromeo/jet-leg}{\texttt{https://github.com/orsoromeo/jet-leg}} \\
(with source code used for Figs. \ref{fig:solveTimesHistogram} and \ref{fig:variableMassActuationRegions}).
\end{itemize}
}
	
\section{Introduction}
{
Most state-of-the-art strategies employed for the online motion planning of legged robots work well only under the assumption that the joint-torque limits do not affect the locomotion capabilities of legged robots. This assumption is due to the computational limitations of modern processors that compel to trade-off between the size of the motion planner's predictive horizon and the accuracy of the considered dynamic model. In order to enlarge the predictive horizon of motion planner for legged robots, therefore, researchers have developed low dimensional (\ie simplified) dynamic models \cite{Full1999} with corresponding stability criteria \cite{Popovic2005}. Such stability criteria focus on avoiding slippage and tip-over \cite{Papadopoulos1996}. They are usually characterized by a \textit{reference point} and a \textit{stability region} in which the reference point has to lie in order to meet the necessary stability condition.
The distance of the considered reference point from the sides of the stability region is used to evaluate the robot's robustness in static and dynamic gaits.}

{Renowned 2D examples of such reference points are the \gls{CoM} projection $\vc{c}_{xy}$, the \gls{zmp} $\vc{z}$ \cite{Vukobratovic2003} and the more recent \gls{icp} $\vc{\xi}$ \cite{Pratt2007}. These 2D reference points represent meaningful descriptors of the system for, respectively, static (neither system's velocity nor accelerations are considered), semi-dynamic (only instantaneous acceleration explicitly considered) and fully dynamic gaits (both instantaneous velocity and acceleration explicitly considered). Furthermore, the \gls{zmp} and the \gls{icp} (both derived by the definition of \emph{linear inverted pendulum} \cite{Kajita2001}) only hold in presence of co-planar contacts. Higher dimensional reference points have been defined to consider also the vertical and angular accelerations such as, respectively, the 3D \textit{virtual repellent point} \cite{Englsberger2013} and the 6D \textit{aggregated centroidal wrench} \cite{Orin2013}.
\begin{figure}
	\centering
	\includegraphics[width=8cm]{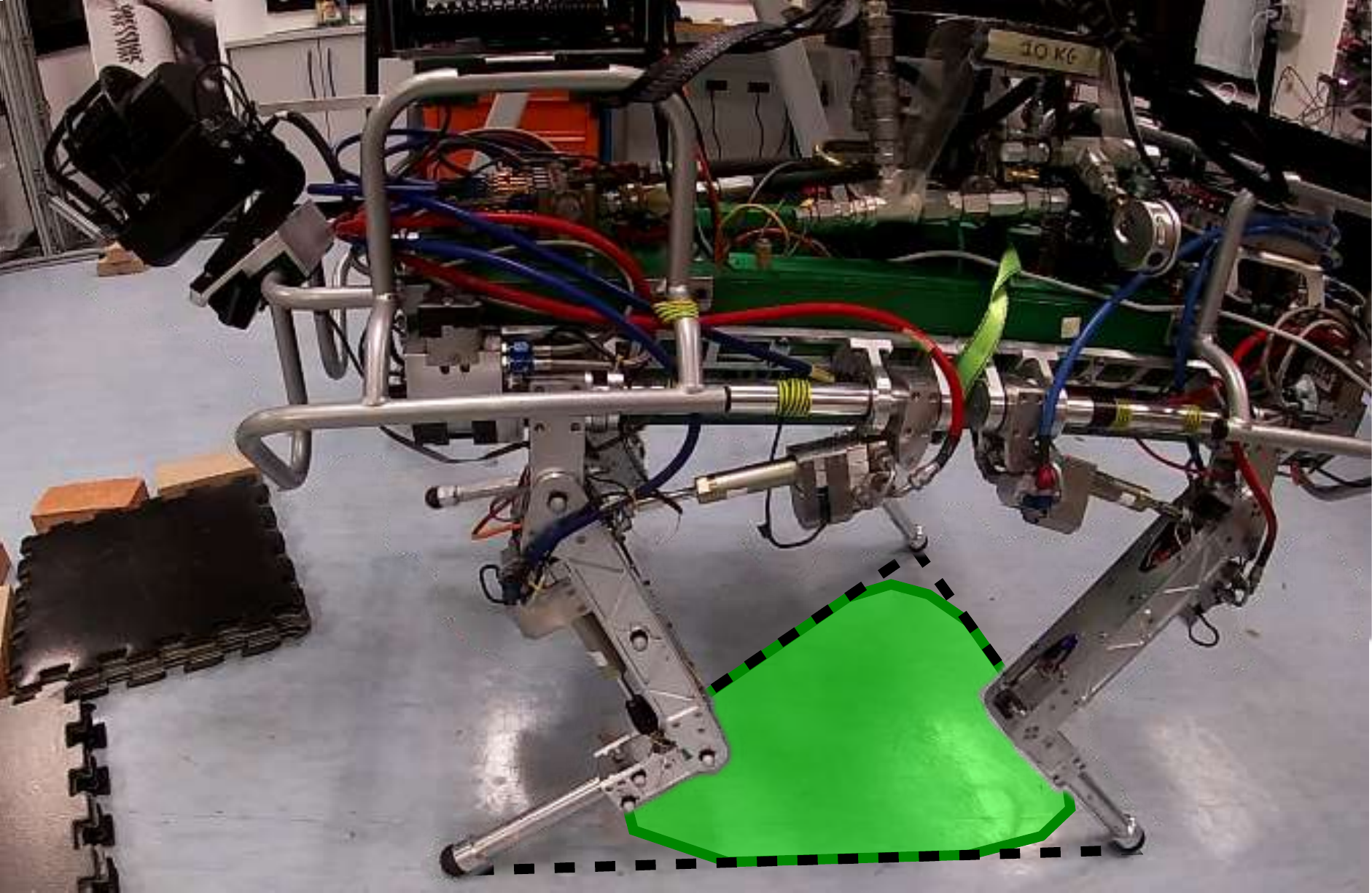}
	\caption[aa]{HyQ robot walk with overlaid classical support region (dashed line) and local Feasible Region (green).}
	\label{fig:frontPic}
\end{figure}}

{As regards the possible definitions of stability region, instead, the \textit{support polygon}, defined as the convex hull of the contact points, is the most widespread. It is always valid in presence of co-planar contacts for both static and dynamic gaits and it can, therefore, be used in combination with the \gls{CoM} projection, the \gls{zmp} or the \gls{icp}. 
It was proved, however, that the support polygon is not valid in multi-contact scenarios and a new definition, called \textit{support region}, was provided \cite{Bretl2008}, \cite{DelPrete2016}. Such definition, however, only holds for the evaluation of static postures on arbitrary terrains, for which reason the support region cannot be combined with \gls{zmp} or \gls{icp}, but only with the \gls{CoM} projection $\vc{c}_{xy}$. A dynamic extension of the support region requires to resort to higher dimensional spaces such as 3D volumes \cite{Audren2017} or 6D sets \cite{Hirukawa2006}, \cite{Saida2003}.}

In all of the above mentioned stability criteria, the reference points capture the main dynamics of the robot's \gls{CoM}, while the stability regions depend on the location of the contact points, the orientation of the surface normals and on the friction coefficients. Extensions have been made in order to include in the lower-dimensional stability analysis also  other feasibility constraints such as the kinematic limits of legged systems \cite{fernbach2018, tonneau2018}. Joint-torque limits, however, have only been mapped into the 6D wrench space \cite{Orsolino2018}, but never into the 2D space of the support region.

In this manuscript we focus on the definition of a 2D stability region that includes the joint-torque limits and, because of the static assumption that it involves, it can be used in combination with the \gls{CoM} projection for the generation of static locomotion for legged robots on complex geometry environments. To the authors' knowledge this is, therefore, the first example of how a two dimensional feasibility criterion can explicitly consider the actuation constraints of the system.

The efficient computation of the proposed \textit{feasible region}, enables the concurrent \textit{online} optimization of actuation-consistent \gls{CoM} trajectories and foothold positions in rough terrains.

Besides its computational efficiency, the \textit{feasible} region can be easily represented in 2D, thus representing an \textit{intuitive} tool for motion planning. This allows us to give a clear and \textit{simple} answer to legitimate questions like: how does the maximal step length change with the increase of the robot's mass? And also: which is the height of the highest step that a robot can step on given its body mass and maximal joint-torque/force capabilities?

\subsection{Contributions}
In this manuscript we attempt to give an answer to the above questions in the following way:
\begin{enumerate}
	\item {We introduce the \textit{feasible region}, a 2D horizontal convex set that includes all the positions of the \gls{CoM} projection that are guaranteed to be statically stable and actuation-consistent. This 2D area provides an \textit{intuitive} yet powerful understanding of the relationship between the task-space locomotion capabilities of a robot and its joint-space actuation limits;}
	\item {We introduce a modified version of the \gls{IP} algorithm, initially proposed in \cite{Bretl2008}, able to efficiently compute the {feasible region} (in about $15ms$); 
	\item We employ the proposed feasible region to formulate a motion planning algorithm for legged robots, able to optimize \textit{online} static trajectories of the \gls{CoM} projection and footholds on arbitrary terrains for predefined step sequences and phase durations.	We report the results of our hardware experiments which show that our motion planner can adapt the footstep locations and the trajectory of the robot's \gls{CoM} to the terrain geometry by replanning in a receding horizon fashion at about \SI[inter-unit-product =\ensuremath{\cdot}]{15}{\hertz}.}
\end{enumerate}

\subsection{Outline}
The two core building blocks of the work developed in this manuscript are the wrench (or \textit{force}) polytopes and the \gls{IP} algorithm which are briefly recalled in Sec. \ref{sec:background}. Using the described elements we formulate the \textit{local actuation region}  $\mathcal{Y}_{a}$ and the \textit{local feasible regions} $\mathcal{Y}_{fa}$ in Sec. \ref{sec:feasibleRegion}. 
%In Section \ref{sec:globalRegion} we then define the \textit{global actuation region} and the \textit{global feasible regions}, two 2D polygons that overcome the local restriction of $\mathcal{Y}_a$ and $\mathcal{Y}_{fa}$. 
The latter, in particular can be seen as a revisited definition of the well known \textit{support region}. 
%Section \ref{sec:feasibleVolume} introduces the possibility of computing 3D feasible volumes that can be used to guarantee that also the height of the robot's \gls{CoM} lies within the set of feasible friction- and actuation-consistent configurations. \\
Section \ref{sec:CoMandFeetPlanning} presents examples of how the feasible region can be employed {to achieve online replanning of \gls{CoM} and feet trajectories} on arbitrary terrains using the height map provided by the robot's exteroceptive sensors. Simulations and experimental results on the HyQ quadruped robot (see Fig. \ref{fig:frontPic}) are finally presented in Sec. \ref{sec:results}. Section \ref{sec:conclusion} draws final conclusions regarding the concepts presented in this manuscript, raises some possible discussion point and anticipates possible future developments.%

\section*{Notation}\label{sec:notation}
List of the most relevant symbols used in this manuscript:
\begin{align*}
	n & \quad \text{Number of actuated joints of the system}\\
			N & \quad \text{\gls{DoFs} of the system}\\
	n_f & \quad \text{Branches of the tree-structured robot}\\
	n_c & \quad \text{Number of contacts}\\
	n_l & \quad \text{Actuated joints of one individual branch}\\
	i & \quad \text{Limb index}\\
	n_i & \quad \text{Degrees of motions of the }\nth{i}\text{ joint.}\\
	m & \quad \text{Dimension of the contact wrench}\\
		\mathbf{c}\in \mathbb{R}^3 & \quad \text{Center of Mass (CoM) position}\\
		\mathbf{p}_i \in \mathbb{R}^3 & \quad \text{End-effector position (hand or foot)}\\
		\mathbf{f}_i\in \mathbb{R}^m & \quad \text{Wrench}\\
		\vc{\tau} \in \Rnum^{n} & \quad \text{Joint-space torques}\\
%	\end{align*}
%	\begin{align*}
		\vc{q} \in SE(3) \times \Rnum^{n} & \quad \text{Point in robot's configuration manifold}\\
		\dot{\vc{q}}_j \in \Rnum^{6+n} & \quad \text{Joint-space velocity}\\
	\vc{q}_i \in \Rnum^{n_l} & \quad \text{Joints configuration of one single branch}\\
		s \in \Rnum^{6+n} & \quad \text{Vector of generalized velocities}\\
		\mathbf{I}_3 \in \Rnum^{3 \times 3} & \quad \text{Identity matrix}\\
		\mathcal{Y}_f & \quad \text{Support (or \textit{friction}) region}\\
		\mathcal{Y}_a & \quad \text{Actuation region}\\
		\mathcal{Y}_{fa} & \quad \text{Feasible region}  \end{align*}
  {Acronyms:
  \begin{align*}
  	\text{CWC} & \quad \text{Contact Wrench Cone}\\
   	\text{AWP} & \quad \text{Actuation Wrench Polytope}\\
	\text{FWP} & \quad \text{Feasible Wrench Polytope}\\
	\text{IP} & \quad \text{Iterative Projection}\\
  \end{align*}}
        
\section{Background}\label{sec:background}

\subsection{Wrench Polytopes for Fixed Base Systems}\label{sec:forcepolygons}
Actuator force/torque limits and their consequences on the overall performance in the task space have been analyzed for decades in the field on mechanical industrial manipulators \cite{Melchiorri1993, Chiacchio1997, Chiacchio1998} and, more recently, also on cable driven parallel robots \cite{CruzRuiz2015} and robotics hands \cite{Krug2016}.\\ 
Wrench ellipsoids (or hyperellipsoids) have been identified as useful tools to assess the control authority at the end-effector of serial mechanical chains. {Being always obtained from a hypersphere of unit radius, such ellipsoids} do not hold any information relative to the absolute magnitude of the wrench that a mechanical chain can exert. They can be obtained as a consequence of the kinetic energy theorem (or work-energy theorem) that states that the work done by all forces acting on a particle equals the rate of change in the particle's kinetic energy \cite[p. 148]{Siciliano2008}. This leads to the following:
\begin{equation}
\vc{\tau} = \vc{J}(\vc{q})^T\vc{w}
\label{eq:transposedJac}
\end{equation}
which represents a static relationship between the generalized task-space wrenches $\vc{w} \in \Rnum^m$ and the generalized joint-space forces $\vc{\tau} \in \Rnum^n$. The matrix $\vc{J}(\vc{q}) \in \Rnum^{m \times n}$ is the end-effector Jacobian.
If we consider  \eqref{eq:transposedJac} in combination with a unit hypersphere $\mathcal{S}_{\tau}$ in the joint-torque space:
\begin{equation}\label{eq:forceEllipsoid}
\mathcal{S}_{\tau} = \Big\{\vc{\tau} \in\Rnum^n \quad | \quad \vc{\tau}^T \vc{\tau} \leq 1 \Big\}
\end{equation}
we can then obtain a new set (the  wrench ellipsoid) $\mathcal{E}_{\vc{w}}$ that describes how $\mathcal{S}_{\tau}$ is mapped into the task-space:
\begin{equation}\label{eq:forcepolytope}
\mathcal{E}_{\vc{w}} = \Big\{\vc{w} \in\Rnum^m \quad | \quad \vc{w}^T \vc{J} \vc{J}^T \vc{w} \leq 1 \Big\}
\end{equation}
By definition, the force ellipsoid $\mathcal{E}_{\vc{w}}$ represents the pre-image of the unit hypersphere $\mathcal{S}_{\tau}$ in the joint space under the mapping given by $\vc{J}(\vc{q})^T$. The lengths of the semiaxes of $\mathcal{E}_{\vc{w}}$ correspond to the singular values of the Moore-Penrose pseudoinverse of $\vc{J}(\vc{q})$ \cite[p.285]{Siciliano2007}. The ratio between the greatest and the smallest eigenvalue of $\vc{J}(\vc{q})$ is, therefore, used as a measure of anisotropy of the ellipsoid and of the force amplification properties of the mechanical chain. 

In a similar fashion, further exploiting ~\eqref{eq:transposedJac}, we can then also analyze the pre-image of the joint-torque hypercube $\mathcal{Z}_{\tau}$, \ie the set of all joint-torques $\vc{\tau}$ comprised within the manipulator's actuator limits $\vc{\tau}^{lim}$:
\begin{equation}\label{eq:jointTorquePolytope}
\mathcal{Z}_{\tau} = \Big\{\vc{\tau} \in \Rnum^n \quad | \quad \vc{\underline{\tau}} \leq \vc{\tau} \leq \vc{\overline{\tau}} \Big\}
\end{equation}
The vectors $ \vc{\underline{\tau}} = - \vc{\tau}^{lim} \in \Rnum^{n}$ and $ \vc{\overline{\tau}} = \vc{\tau}^{lim} \in \Rnum^{n}$ contain in their elements the hardware-dependent lower and upper bounds of the values that limit the generalized joint-torque vector $\vc{\tau}$\footnote{{For electric and rotary hydraulic actuators, provided that the actuator's friction is properly modelled, the dependence of the torque limits on the joint position is negligible. The torque limits of electric actuators also depend on the joint velocity; it was shown, however, that such dependency is often negligible compared to effects of other phenomena such as temperature \cite{Vaillant2014}. In this work, moreover, we focus on static locomotion for which reason the torque limit $\tau_{lim}$ can be identified with the maximum static \textit{stall-torque}.}}. 

\begin{figure}
		\centering
		\begin{subfigure}{7cm}
			\includegraphics[width=1\textwidth]{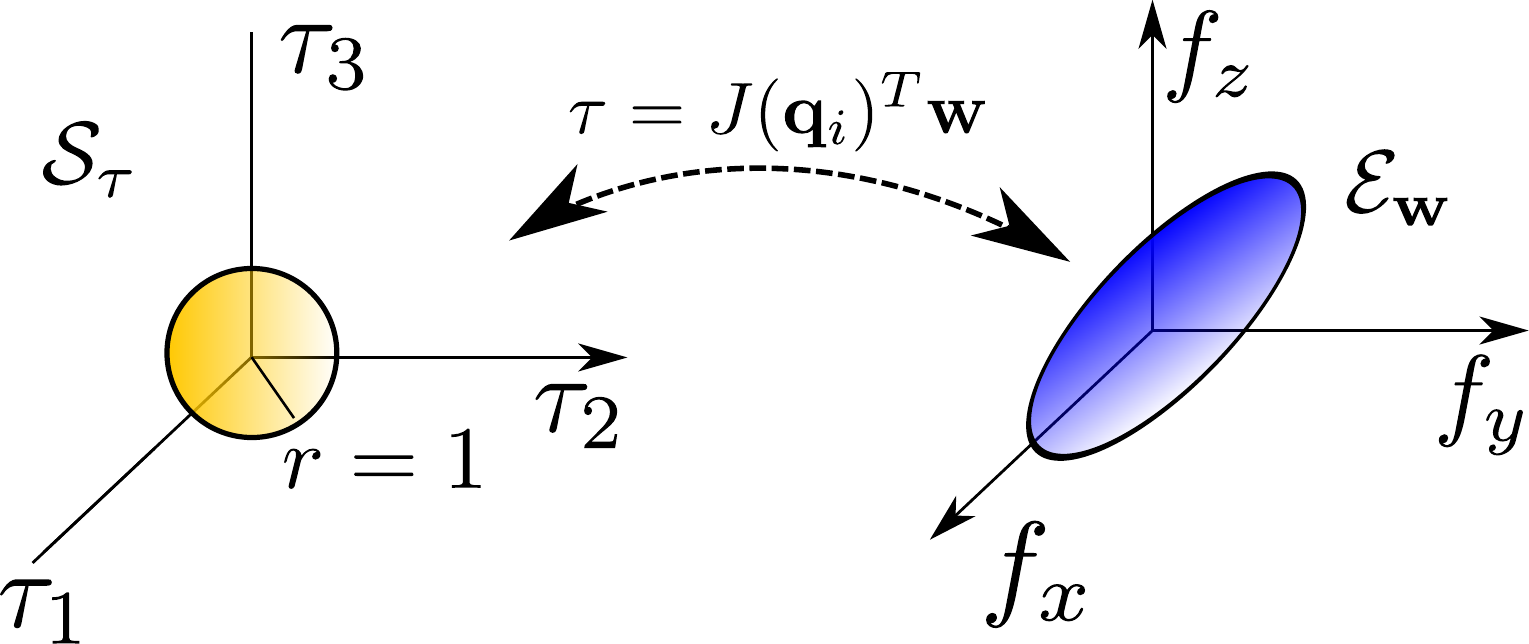}
			\caption{Unit sphere mapping into a wrench ellipsoid}
			\label{fig:staticTorquesMappingSphere}
		\end{subfigure}
		~
		\begin{subfigure}{7cm}
			\includegraphics[width=1\textwidth]{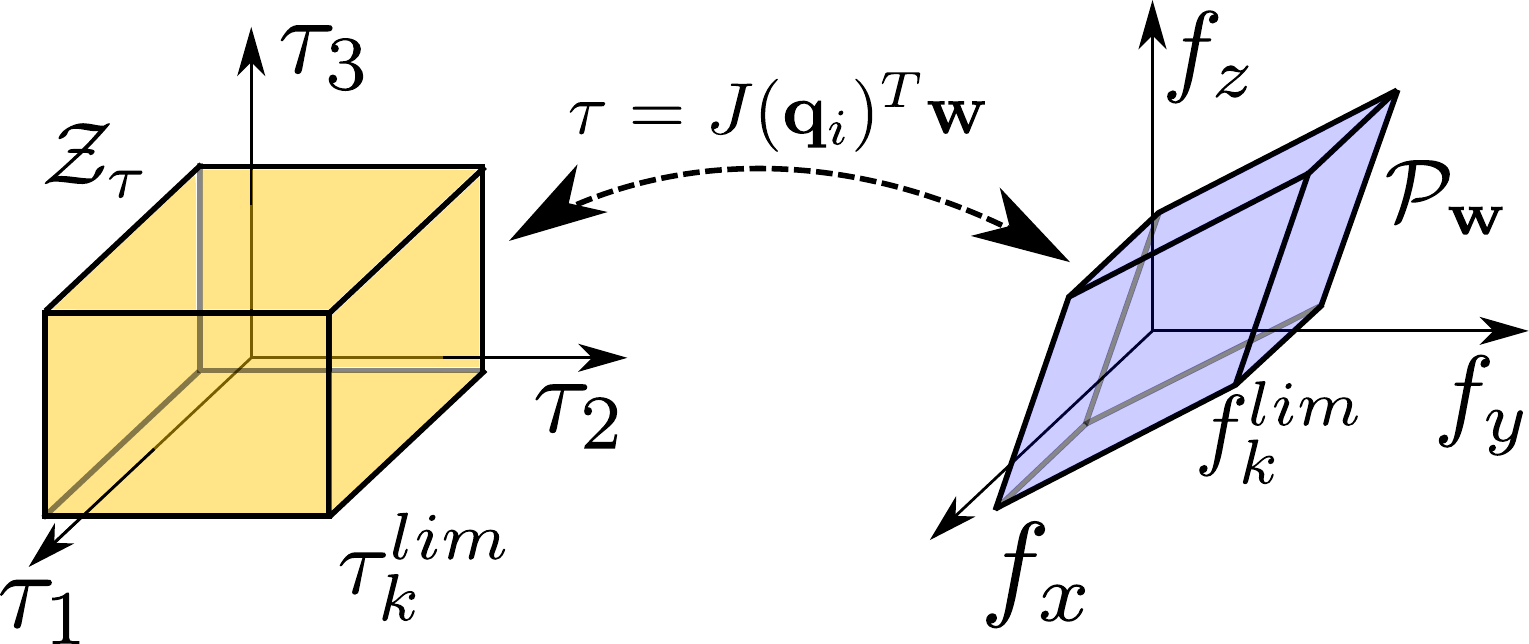}
			\caption{Zonotope mapping into a wrench polytope}
			\label{fig:staticTorquesMappingPolygon}
		\end{subfigure}
		
		\captionsetup{singlelinecheck=off}
		\caption[Joint torques to contact forces mapping]{The mapping between joint-space torques and the task-space forces at the end-effector. In this example, the dimension of joint-torque space $dim(\mathcal{Z}_{\vc{\tau}}) = dim(\mathcal{S}_{\vc{\tau}}) = n = 3$ is equal to the dimension of the manifold of the contact forces $dim(\mathcal{P}_{\vc{w}}) = dim(\mathcal{E}_{\vc{w}}) = m = 3$. The index $i = 0, \dots , n_l$ represents the limb's index while $k = 0, \dots , 2^n$ represents the vertices' index.}
		\label{fig:staticTorquesMapping}
	\end{figure}

The hypercube $\mathcal{Z}_{\tau}$ can be seen also as a system of $2n$ linear inequalities that constrain joint-torques~\cite{Chiacchio1998} (see Fig. \ref{fig:staticTorquesMapping}). The notation used in ~\eqref{eq:jointTorquePolytope} assumes {symmetric joint-torque limits\footnote{{We assume that the maximum joint-torque $\vc{\overline{\tau}}$ has the same amplitude of the minimum joint-torque $\vc{\underline{\tau}}$ and opposite sign ($\vc{\overline{\tau}} = \tau^{lim}$ and $\vc{\underline{\tau}} = -\tau^{lim}$).}} and, in this case, $\mathcal{Z}_{\tau}$ is a zonotope centered at the origin (see Appendix~\ref{appendix:boundedPolygonsAndCones}). In all the other cases the hypercube $\mathcal{Z}_{\tau}$ will still represent a zonotope but its center will not correspond to the origin of joint-torque space.}
	
The task-space wrench polytope $\mathcal{P}_{\vc{w}}$\footnote{Also commonly called \emph{force polytope} \cite{Chiacchio1997}}, pre-image of $\mathcal{Z}_{\tau}$, can be written as follows (also see Fig. \ref{fig:staticTorquesMapping}):
\begin{equation}\label{eq:jointForcePolytope}
\mathcal{P}_{\vc{w}} = \Big\{ \vc{w} \in \Rnum^m \quad | \quad \vc{\underline{\tau}} \leq \vc{J}(\vc{q})^T\vc{w} \leq \vc{\overline{\tau}} \Big\}
\end{equation}
While the force ellipsoid $\mathcal{E}_{\vc{w}}$ can be used as a qualitative metric of the robot's force amplification capabilities, the wrench polytope $\mathcal{P}_{\vc{w}}$ also includes a quantitative information about the maximum and minimum amplitude of the wrench that the robot can perform at the end-effector.

Wrench ellipsoids and wrench polytopes have been originally introduced for fixed-base non-redundant serial mechanical chains with $m = n$ where $n$ is the number of actuated joints (dimension of generalized coordinates) and $m$ is the dimension of the end-effector force (or, equivalently, the degree of constraint at the contact). In such cases the Jacobian $\vc{J}(\vc{q})$ is thus {a square matrix} and, except for singular configurations, its transpose $\vc{J}^T$ can be inverted to obtain the {vertex representation} of the wrench polytope $\mathcal{P}_{\vc{w}}$:
\begin{equation}
\vc{w}^{lim} = \vc{J}(\vc{q})^{-T} \vc{\tau}^{lim}
\end{equation}
where $\vc{\tau}^{lim} \in \Rnum^n$ is a vertex of $\mathcal{Z}_{\vc{\tau}}$ and $\vc{w}^{lim} \in \Rnum^m$ is a vertex of $\mathcal{P}_{\vc{w}}$. This is a suitable condition in which a one-to-one relation between joint-space torques and task-space wrenches exists. In the case of an arm with 3 \gls{DoFs} ($n = 3$), for example, a violation of one joint-torque limit will correspond to a point on a facet of $\mathcal{Z}_{\tau}$ and also to another point on a facet of the task-space polytope $\mathcal{P}_{\vc{w}}$. Similarly, a violation of two (or three) joint-torque limits will correspond to a point on an edge (or a vertex) of the $\mathcal{Z}_{\tau}$ and also to another point on an edge (or a vertex) of the task-space polytope $\mathcal{P}_{\vc{w}}$. See Appendix \ref{appendix:boundedPolygonsAndCones} for the definitions of facets, edges and vertices for $n > 3$.

For a more detailed explanation about the effect of gravity on force ellipsoids and on other possible definitions of wrench polytopes for fixed base systems in dynamic conditions please refer to \cite{Chiacchio1992, Chiacchio1998, Yoshikawa1985}.

\subsection{Wrench Polytopes for Floating Base Systems}
{In this section we illustrate the procedure to compute the \textit{dynamic  wrench polytopes} $\mathcal{A}$ \ie the set of feasible contact wrenches that a tree-structured, floating base robot can perform at its contact points with the environment while moving \cite{Chiacchio1997}}. For this, let us consider the \gls{EoM} of a floating-base robot\footnote{We consider the floating-base robot composed by $n_f$ branches (\eg number of feet and/or hands), with $n_c$ of them in contact
with the environment and, each of them having a number $n_l$ of actuated \gls{DoFs}. Therefore, $n = \sum_{k=1}^{n_f} n_{l}^{k}$ represents the total number of actuated joints.}:
%
%\begin{equation}\label{eq:floatingBaseDynamics}
%\vc{M}(\vc{q})\vc{\ddot{q}} + \vc{c}(\vc{q}, \vc{\dot{q}}) +
%\vc{g}(\vc{q}) = \vc{B \tau} + \vc{J}_s^T(\vc{q}) \vc{f}
%\end{equation}
%
\begin{equation}\label{eq:floatingBaseDynamics}
\vc{M}(\vc{q})\vc{\dot{s}} + \vc{C}(\vc{q}, \vc{{s}}) +
\vc{g}(\vc{q}) = \vc{S \tau} + \vc{T}(\vc{q})^T \vc{f}
\end{equation}
where $\vc{q}=\mat{\vc{q}_b^T & \vc{q}_j^T}^T \in SE(3) \times \Rnum^n$ is the vector of generalized coordinates of the floating-base system, composed of the pose of the floating base $\vc{q}_b \in SE(3)$ and of the coordinates $\vc{q}_j\in\Rnum^n$ describing the positions of the $n$ actuated
joints. 
The vector $\vc{s}=\mat{\vc{\nu}_b^T & \vc{\dot{q}}_j^T}^T \in \Rnum^{6+n}$ is the vector of the generalized velocities, 
$\vc{\tau}\in\Rnum^n$ is the vector of actuated joint-torques while
$\vc{C}(\vc{q}, \vc{s})$ and $\vc{g}(\vc{q}) \in \Rnum^{6+n}$ are the
{Centripetal}/Coriolis and gravity terms, respectively. The matrix $\vc{M}(\vc{q}) \in \Rnum^{(n+6)\times(n+6)}$ is the joint-space inertia matrix, $\vc{S} \in \Rnum^{(6+n)
\times n}$ is the {selector matrix whose rows corresponding to the actuated joints are set to ones and whose rows corresponding to the unactuated joints are set to zeros.} $\vc{f}\in \Rnum^{m n_c}$ is the vector of contact forces\footnote{Note that quadruped robots have nearly point feet, henceforth we thus consider pure forces acting at contact points and no contact torque ($m = 3$).} that are mapped into joint torques through the stack of Jacobians $\vc{T}(\vc{q})\in\Rnum^{m n_c\times(6+n)}$. If we split ~\eqref{eq:floatingBaseDynamics} into its
unactuated and actuated parts, we get:
\begin{equation}
\underbrace{\mat{\vc{M}_b & \vc{M}_{bj} \\ \vc{M}_{bj}^T & \vc{M}_j}}_{\vc{M}(\vc{q})}
\underbrace{\mat{\vc{\dot{\nu}_b} \\ \vc{\ddot{q}}_j} }_{\vc{\dot{s}}} +
\underbrace{\mat{\vc{c}_b \\ \vc{c}_j} }_{\vc{C}(\vc{q},\vc{s})} +
\underbrace{\mat{\vc{g}_b \\ \vc{g}_j} }_{\vc{g}(\vc{q})} =
\underbrace{\mat{\vc{0}_{6\times n} \\ \vc{I}_{n\times n}}}_{\vc{B}}\vc{\tau} +
\underbrace{\mat{\vc{J}_{b}^T \\ \vc{J}_{q}^T}}_{\vc{T}(\vc{q})^T}\vc{f}.
\label{eq:FBdynEq}
\end{equation}
{The vector $\vc{q}_j$ results from the concatenation of the joint positions of all the separate branches $\vc{q}_1, \dots \vc{q}_{n_c}$. The bottom block of $\vc{T}(\vc{q})$ (bottom $n$ rows), corresponding to the actuated part $\vc{J}_{q} \in \Rnum^{(m n_c) \times n}$, is thus block diagonal and it can map joint-torques into contact forces for each leg
\textit{individually}:}
\begin{equation}
	\vc{J}_{q} = \diag(\vc{J}(\vc{q}_1), \dots \vc{J}(\vc{q}_{n_c}))
\end{equation}
where $\vc{J}(\vc{q}_i)$ with $i = 1, \dots n_c$ are the Jacobians of the limbs in contact with the ground.
Omitting the {upper block (first $6$ rows) of ~\eqref{eq:FBdynEq}} is convenient because it avoids the coupling term $\vc{J}_{b}$ and one  wrench polytope can then be computed for each individual limb.

On a similar line, as the dynamic manipulability polytope $\mathcal{P}_{\vc{w}}$ defined in \cite{Melchiorri1993}, we can now define a quantity that we call \textit{dynamic wrench polytope} $\mathcal{A}_i$ {as follows:}	

\textbf{Definition:} for each individual $\nth{i}$ branch of the considered tree-structured robot, $\mathcal{A}_i$ corresponds to the set of all contact forces $\vc{f}_i \in \Rnum^m$ that satisfies the bottom row of ~\eqref{eq:FBdynEq} for all the joint-torques $\vc{\tau}_i \in \Rnum^{n_l}$ belonging to $\mathcal{Z}_{\tau}$:
\begin{equation}\label{eq:dynamicForcepolytope}
\begin{aligned}
\mathcal{A}_{i} = \Big\{&  \vc{f}_i \in \Rnum^m  | \exists \vc{\tau}_i \in \Rnum^{n_l} s.t. \quad  \vc{M}_{bi}^T	\vc{\dot{\nu}} + \vc{M}_i \vc{\ddot{q}}_i +  \vc{c}(\vc{q}_i, \dot{\vc{q}}_i) + \\ & \vc{g}(\vc{q}_i) = 
\vc{\tau}_i + \vc{J}(\vc{q}_i)^T \vc{f}_i, \quad \vc{\underline{\tau}}_i \leq \vc{\tau}_i \leq \vc{\overline{\tau}}_i \Big\}
\end{aligned}
\end{equation}
where $i = 1, \dots, n_c$ is the contact index and $n_c$ is the number of active contacts between the robot and the environment. The vectors $\vc{q}_i \in \Rnum^{n_l}$ and $\vc{\tau}_i \in \Rnum^{n_l}$ represent the joint-space position and torque of only those joints that belong to the $\nth{i}$ limb while $n_l$ represents the number of actuated \gls{DoFs} of that limb. If $m = 3$ then the contact wrench $\vc{f}_i \in \Rnum^m$ consists of pure forces while if $m=6$ then a non-zero contact torque is also present. For a partial list of the main symbols employed in this paper and their meaning please refer to the Notation Section \ref{sec:notation}.

\begin{figure}
		\centering
		\includegraphics[width=7cm]{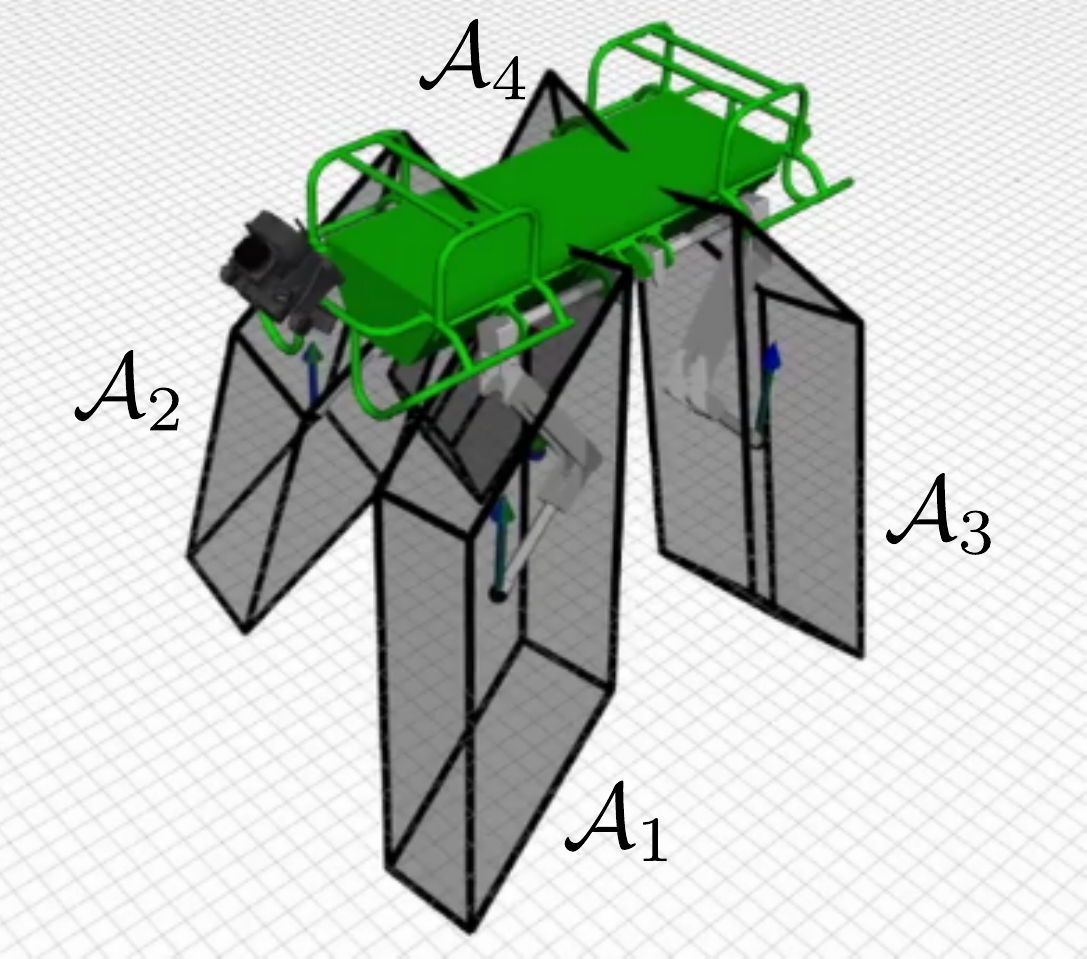}
		\caption[Actuation polytopes for HyQ]{Representation of the wrench polytopes $\mathcal{A}_i$ on the feet of the HyQ robot ($i = 0, \dots, n_c$ is the leg index). {The wrench polytopes $\mathcal{A}_{i}$ should be intersected with the friction cones $\mathcal{F}_{i}$ to enforce unilaterality of the contact forces and to avoid slippage.}}
		\label{fig:actuationAndFriction}
	\end{figure}

In Fig. \ref{fig:actuationAndFriction} an example of dynamic wrench polytope is drawn {superimposed on} the \gls{HyQ} robot: each limb of this robot has three actuated \gls{DoFs} ($n_l = 3$) and each foot can be approximated as a point contact ($m = 3$). $\mathcal{A}_{i}$ is then a polytope of $2\cdot3 = 6$ facets and $2^3 = 8$ vertices (the mapping of a cube in the joint space).
Equation~\eqref{eq:dynamicForcepolytope} purposely omits the first line (six equations) of ~\eqref{eq:FBdynEq} referring to the unactuated floating base. This corresponds to neglecting the role of the base link as a coupling among the limbs of the tree-structured robot. \

The advantage of computing separate individual wrench polytopes $\mathcal{A}_i$ for each limb of the robot is that we can then treat {the volume of each polytope as an actuation capability} measure of the corresponding limb \cite{Samy2017, Boussema2019}.

As a final consideration, we can observe that in static conditions ($\dot{\vc{q}}_i = \ddot{\vc{q}}_i = 0$) ~\eqref{eq:dynamicForcepolytope} can be written as: 
\begin{equation}\label{eq:staticForcepolytope}
\begin{aligned}
\mathcal{A}_i = \Big\{ \vc{f}_i \in \Rnum^m \quad | \quad &\exists \vc{\tau}_i \in \Rnum^{n_l} s.t. \\ \vc{g}(\vc{q}_i) = \vc{\tau}_i + \vc{J}(\vc{q}_i)^T \vc{f}_i, & \quad \vc{\underline{\tau}}_i \leq \vc{\tau}_i \leq \vc{\overline{\tau}}_i \Big\}
\end{aligned}
\end{equation}
The term $\vc{g}(\vc{q}_i)$ represents the effect of gravity acting on the individual limb $i = 0, \dots, n_c$. From a geometrical point of view $\vc{g}(\vc{q}_i)$ can also be seen as an offset term that translates the polytope $\mathcal{A}_i$ in the same direction of the gravity vector, \ie towards the negative side of the $f_z$ direction of the wrench space defined by the axes ($f_x, f_y, f_z, \tau_x, \tau_y, \tau_z$). For a predefined set of torque limits $\vc{\tau}_i^{lim}$ an increase in the legs' mass and, as a consequence, a large offset term $\vc{g}(\vc{q}_i)$, will cause a decrease in the set of feasible positive contact forces. 
%Just as the \gls{cwc}, the \gls{AWP} and the \gls{FWP} are valid for arbitrary contacts (not limited to flat terrains) and for dynamic motions with non-zero linear and angular momentum rates. \\
%The computation time of six-dimensional bounded polytopes (\gls{AWP} and \gls{FWP}) increases considerably with respect to the case where only six-dimensional convex cones are involved\footnote{Convex cones hold the convenient property that their Minkowsky sum corresponds to their convex hull, thus making the computation of the \gls{cwc} considerably faster than the \gls{FWP} computation (see Appendix \ref{appendix:minkowskiSums}).} (\eg \gls{cwc}). In the case of a quadruped robot, for example, the \gls{FWP} constraints can only be computed at about $10$~Hz in a triple stance phase and about $3$~Hz during a quadruple stance phase\cite{Orsolino2018}. Such computational performances allow \textit{online} motion planning to be achieved only under the assumption that the robot's configuration is not too far from the nominal configuration used to obtain the \gls{FWP} limits. These restrictions, imposed by computational performances of modern processors, limit the effectiveness of wrench-based actuation-aware online planning.\\

\subsection{The Iterative Projection Algorithm}\label{sec:IPalgorithm}
The Iterative Projection (IP) algorithm is a method introduced by Bretl \etal \cite{Bretl2008} for the computation of support regions for articulated robots having multiple contacts with the environment in arbitrary locations, having arbitrary surface normals and friction coefficients. In \cite{Bretl2008}, the support region is defined as \textit{the horizontal cross section of the convex cylinder that represents the set of \gls{CoM} positions at which contact forces exist that compensate for gravity without causing slip (for given foot placements with associated friction models)}.\\ 
The \gls{IP} belongs to a family of cutting-plane methods~\cite{Kelley1960} that allow to approximate a target convex set $\mathcal{Y}$ up to a predefined tolerance value. The tuning of this tolerance allows to conveniently adjust the computational performance of the algorithm: it enables a rough but fast reconstruction of the target set for high tolerance values. On the other hand, it also enables a precise set reconstruction with longer solve times when the tolerance value is low. 

Bretl \etal~\cite{Bretl2008} have applied this algorithm to the field of legged locomotion with the goal to reconstruct the 2D friction-consistent support region $\mathcal{Y}_f$ for the \gls{CoM} in static equilibrium. This algorithm was also applied in related works to compute multi-contact ZMP support areas~\cite{Caron2017b} or time-optimal trajectory timings~\cite{Hauser2014, Caron2017c}. In order to reduce the confusion with other similar regions that will be defined in the upcoming Sections, in the remainder of this manuscript the support region for the \gls{CoM} in static equilibrium will be referred to simply as the \textit{friction region}.

Algorithm \ref{alg:iterativeProjectionBretl} reports the procedure presented in \cite{Bretl2008}; one can notice that the algorithm recursively solves a {\gls{lp}} that maximizes the horizontal position of the \gls{CoM} $\vc{c}_{xy} \in \Rnum^2$ along the direction defined by the unit vector $\vc{a}_j \in \Rnum^2$ ($j$ being in this case the iteration index) while satisfying the friction constraints.

		\begin{algorithm}
			\textbf{Input: }{$\vc{c}_{xy}, \vc{p}_i, \vc{n}_i, \vc{t}_{1,i}, \vc{t}_{2,i}, \mu_i$ for $i = 0, \dots, n_c$ }\;
			\KwResult{friction region $\mathcal{Y}_f$}
			initialization: $\mathcal{Y}_{outer}$ and $\mathcal{Y}_{inner}$\;
			\While{$area(\mathcal{Y}_{outer}) - area(\mathcal{Y}_{inner}) > \epsilon$}{
				I) compute the edges of $\mathcal{Y}_{inner}$\;
				II) pick the edge cutting off the largest fraction of $\mathcal{Y}_{outer}$\;
				III) solve the LP:\:
				\hspace{4cm}$\begin{aligned}
				\hspace{1.5cm} \max_{\vc{c}_{xy}, \vc{f}} \quad &  \vc{a}_j^T \vc{c}_{xy}\\
				\text{such that:}\\
				(\text{III.a}) & \quad \vc{A}_1\vc{f} + \vc{A}_2\vc{c}_{xy} = \vc{u}\\
				(\text{III.b}) &	\quad    \vc{B}\vc{f} \leq \vc{0} 
				\end{aligned}$
				
				\:
				IV) update the outer approximation $\mathcal{Y}_{outer}$\;
				V) update the inner approximation $\mathcal{Y}_{inner}$\;
			}
			
			\caption{Bretl and Lall's IP algorithm}
			\label{alg:iterativeProjectionBretl}
		\end{algorithm}
	
{The algorithm considers the projection of the robot's \gls{CoM} $\vc{c}_{xy}$, the mass $m$, a set of $n_c$ contacts $\vc{p}_i \in \Rnum^3$ with corresponding surface normals $\vc{n}_i \in \Rnum^3$, tangent vectors $\vc{t}_{1,i}, \vc{t}_{2,i} \in \Rnum^3$ and friction coefficients $ \mu_i \in \Rnum$ (for $i = 0, \dots, n_c$).} The constraint (III.a) enforces the static equilibrium of the forces and moments acting on the robot due to gravity $\vc{g} \in \Rnum^3$ and to the contact forces $\vc{f} = [\vc{f}_1^T, \dots, \vc{f}_{n_c}^T]^T \in \Rnum^{m{n_c}}$. The matrix $\vc{A}_1  \in \Rnum^{6 \times m n_c}$ represents the grasp matrix of the set of point contacts\footnote{$[\cdot]\times$ represents the skew-symmetric operator associated to the cross product. $m = 6$ for generic contacts and $m = 3$ for point contacts.}, $\vc{A}_2 \in \Rnum^{6 \times 2}$ computes the $x,y$ angular components $\tau_\mathcal{O}^x$ and $\tau_\mathcal{O}^y$ of the wrench generated by the action of gravity on the \gls{CoM}  $\vc{c}$ of the robot expressed with respect to a fixed frame $\mathcal{O}$.  {Constraint (III.b) ensures that the friction constraint is satisfied through the matrix $\vc{B} \in \Rnum^{4 n_c \times 3 n_c} $. Note that this represents the friction pyramids as an approximation of the more precise friction cones~\cite{Trinkle1997}. Such friction cones could be enforced in (III.b) thus transforming the \gls{lp} of step (III) into a \gls{SOCP} with a negligible computational overhead.
	\begin{equation}\label{eq:IPmatrices}
	\begin{aligned}
	&\vc{A}_1 = \mat{\bar{\vc{A}}_1 & \dots & \bar{\vc{A}}_{n_c}} \in \Rnum^{6\times (m {n_c})}, \\
		& 	\vc{A}_2 = \mat{\mathbf{0} \\
		-m \vc{g} \times \vc{P}^T} \in \Rnum^{6\times 2}, \quad \vc{P} = \mat{1 & 0 & 0\\
		0 & 1 & 0}\\
	&	\vc{u} = \mat{ - m \vc{g} \\ \vc{0}},
	\quad \vc{b}_i = \mat{
		(\vc{t}_{1,i} - \mu_i \vc{n}_i)^T\\ (\vc{t}_{2,i} - \mu_i \vc{n}_i)^T \\ -(\vc{t}_{1,i} + \mu_i \vc{n}_i)^T \\ - (\vc{t}_{2,i} + \mu_i \vc{n}_i)^T} \in \Rnum^{4 \times 3} \\	
	 &\vc{B} = \diag(\vc{b}_1, \dots, \vc{b}_{n_c}) \in \Rnum^{4 n_c \times 3 n_c}, \quad \vc{g} = [0,0,-g]^T
	\end{aligned}
	\end{equation}}
	where $\vc{P} \in \Rnum^{2 \times 3}$ is a selection matrix that selects the $x,y$ components of the \gls{CoM} and $\bar{\vc{A}}_i$ is a transformation matrix such that:	
	\begin{equation}
	\bar{\vc{A}}_i = \left\{ \begin{tabular}{ll}
	\vspace{0.1cm}
		 \mat{\mathbf{I}_3 \\
		 			[\vc{p}_i] \times} $ \in \Rnum^{6\times 3{n_c}}$ & \text{if} \quad $m = 3$\\ 
		 \mat{\mathbf{I}_3 & \mathbf{0}_3\\
		 	 			[\vc{p}_i] \times & \mathbf{I}_3}  $\in \Rnum^{6\times 6{n_c}}$ & \text{if} \quad $m = 6$ \\
		\end{tabular} \right.
	\end{equation}
	With the above definitions, the \textit{friction region} $\mathcal{Y}_f$ is defined as the set of horizontal \gls{CoM} coordinates $\vc{c}_{xy} \in \Rnum^2$ for which there exists a set of contact forces that respects both static equilibrium and friction constraints:
	\begin{equation}\label{eq:supportRegionBretl}
	\mathcal{Y}_f = \Big\{\vc{c}_{xy} \in\Rnum^2 \quad | \quad \exists \vc{f}\in\Rnum^{m n_c} \st (\vc{c}_{xy}, \vc{f}) \in \mathcal{C} \Big\}
	\end{equation}
	where:
	\begin{equation}\label{eq:feasibleSetBretl}
	\begin{aligned}
	\mathcal{C} = \Big\{\vc{f}\in\Rnum^{m n_c}, \quad & \vc{c}_{xy} \in\Rnum^2 \quad | \quad \vc{A}_1\vc{f} + \vc{A}_2\vc{c}_{xy} = \vc{u},\\
	& \vc{B}\vc{f} \leq \vc{0}  \Big\}
	\end{aligned}
	\vspace{+0.3cm}
	\end{equation}

	\noindent%
	\begin{minipage}{\linewidth}% to keep image and caption on one page
		\makebox[\linewidth]{%        to center the image
			\includegraphics[keepaspectratio=true,scale=0.45]{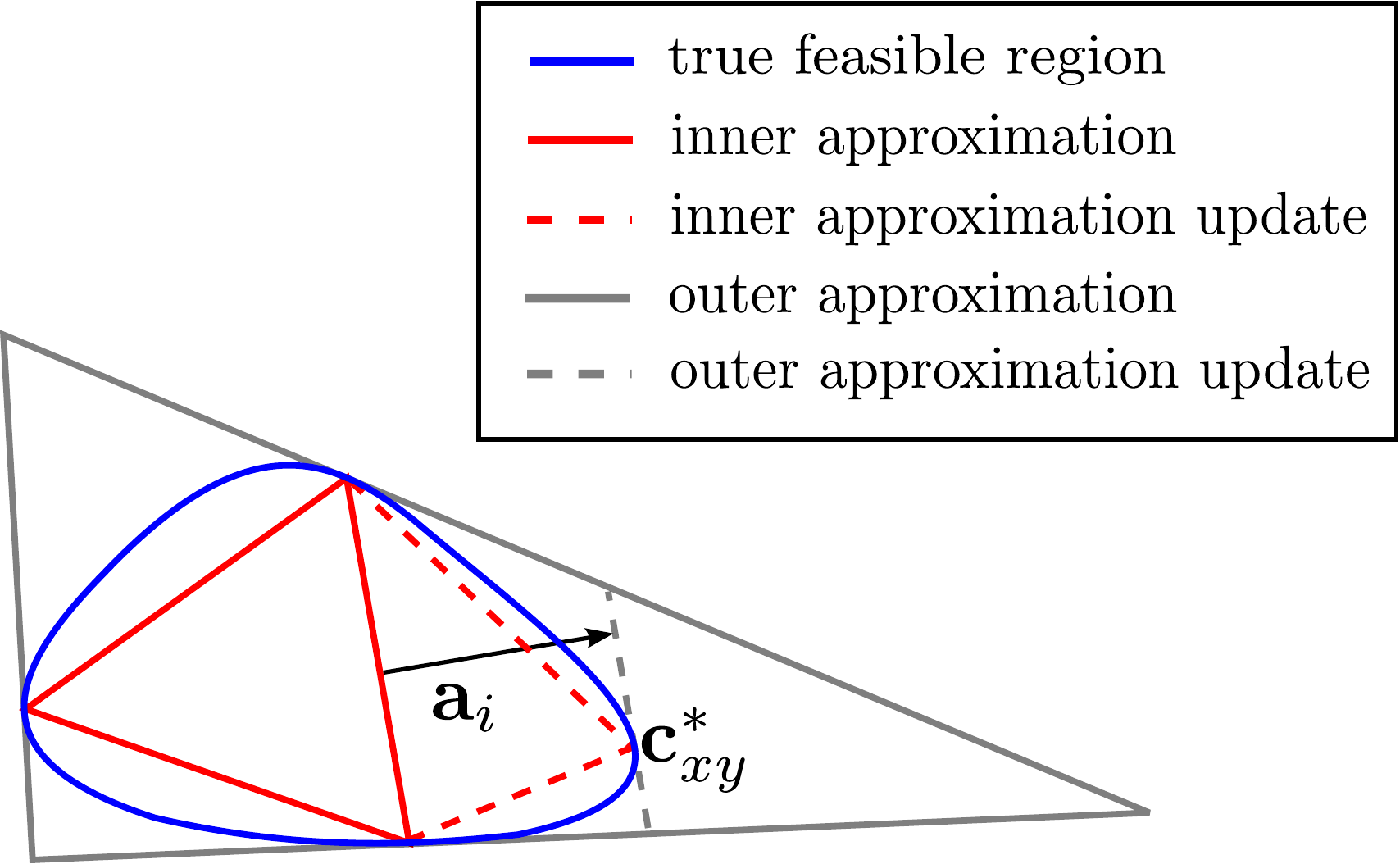}}
		\captionof{figure}[Bretl and Lall's iterative projection algorithm]{Single step of Bretl and Lall's IP algorithm. The recursive \gls{lp}s explore and expands the feasible region in the direction where the difference between the inner and the outer approximation is the largest.}\label{fig:bretlIterativeProjection}%      only if needed  
	\end{minipage}
	\vspace{.3cm}

Fig. \ref{fig:bretlIterativeProjection} shows the process to compute one iteration of the \gls{IP} algorithm reported in Alg. \ref{alg:iterativeProjectionBretl}. As it can be seen in step III, the \gls{IP} does not only maximize the horizontal \gls{CoM} projection $\vc{c}_{xy}$ along the direction $\mathbf{a}_i \in \Rnum^2$, but it also finds a feasible set of contact forces $\vc{f}$ that fulfills static equilibrium and friction cone constraints (constraints III.a and III.b).\\
	Alg. \ref{alg:iterativeProjectionBretl} can also be regarded as a projection of the feasible set $\mathcal{C}$ onto a two-dimensional region whose boundaries represent the limit torques $\tau^x_{\mathcal{O}}, \tau^y_{\mathcal{O}}$, that the robot can exert to balance the effect of gravity acting on its \gls{CoM}. Exploiting the assumption that the only external force acting on the \gls{CoM} is gravity we then get a one-to-one mapping between the torque components and the corresponding \gls{CoM} ($x,y$) coordinates:
	\begin{equation}
	c_x = \frac{\tau^y_{\mathcal{O}}}{mg}, \quad c_y = -\frac{\tau^x_{\mathcal{O}}}{mg}
	\label{eq:staticMapping}
	\end{equation}
	The friction region, as defined in ~\eqref{eq:supportRegionBretl}, is a 2D convex set but it is not, in general, a linear set (\ie it is not a polygon). The inner and outer approximations $\mathcal{Y}_{inner}$ and $\mathcal{Y}_{outer}$ used to estimate $\mathcal{Y}_f$ are, however, always 2D polygons by construction. For this reason we will therefore refer to $\mathcal{Y}_f$ in the rest of this paper with the term \textit{friction region} rather than \textit{friction polygon}.\\
	Del Prete \etal \cite{DelPrete2016} proposed a Revisited Incremental Projection (IPR) algorithm to test static equilibrium which is shown to be faster than the original \gls{IP} formulation and than other possible techniques such as the Polytope Projection (PP). However the IPR approach is only suitable for convex cones and, therefore, does not fit well with the projection of bounded polytopes that we address in the next Section. 
	
	In the next Section we will see how we modified Alg. \ref{alg:iterativeProjectionBretl} in order to obtain a 2D set that does not only respect the static equilibrium and friction cone constraints, but it also respects the actuation capabilities of the system.

	\section{The Feasible Region}\label{sec:feasibleRegion}
{In this Section we propose the main contribution of this manuscript which is an extension of Alg. \ref{alg:iterativeProjectionBretl} in order to obtain support regions that also consider the robot's joint-torque limits besides the constraints imposed by the friction cones.} 

As opposed to Alg.~\ref{alg:iterativeProjectionBretl}, we include the static wrench polytope $\mathcal{A}_i$ of every individual end-effector in contact with the environment{, as given in ~\eqref{eq:staticForcepolytope}}. The resulting procedure can be found in Alg. \ref{alg:iterativeProjectionWithActuation}.

\begin{algorithm}
				\textbf{Input: }{$\vc{c}_{xy}, \vc{G},\vc{d}, \vc{p}_i, \vc{n}_i, \vc{t}_{1,i}, \vc{t}_{2,i}, \mu_i$ for $i = 0, \dots, n_c$}\;
				\KwResult{feasible region $\mathcal{Y}_{fa}$}
				initialization: $\mathcal{Y}_{outer}$ and $\mathcal{Y}_{inner}$\;
				\While{$area(\mathcal{Y}_{outer}) - area(\mathcal{Y}_{inner}) > \epsilon$}{
					I) compute the edges of $\mathcal{Y}_{inner}$\;
					II) pick the edge cutting off the largest fraction of $\mathcal{Y}_{outer}$\;
					III) solve the LP:\:
					\hspace{4cm}$\begin{aligned}
					\hspace{1.5cm} \max_{\vc{c}_{xy}, \vc{f}} \quad &  \vc{a}_j^T \vc{c}_{xy}\\
					\text{such that}:\\
					(\text{III.a}) & \quad \vc{A}_1\vc{f} + \vc{A}_2\vc{c}_{xy} = \vc{u}\\
					(\text{III.b}) &	\quad    \vc{B}\vc{f} \leq \vc{0} \\
					(\text{III.c}) &	\quad    \vc{G}\vc{f} \leq \vc{d} 
					\end{aligned}$
					
					\:
					IV) update the outer approximation $\mathcal{Y}_{outer}$\;
					V) update the inner approximation $\mathcal{Y}_{inner}$\;
				}
				
				\caption{Actuation and Friction consistent IP.}
				\label{alg:iterativeProjectionWithActuation}
			\end{algorithm}

{In contrast to} the original \gls{IP} algorithm, we also retain the possibility of exerting contact torques at the end-effectors; as a consequence we define each individual contact wrench $\vc{f}_i \in \Rnum^m$ where $m = 3$ if the considered end-effector is perturbed by a pure force and $m=6$ if, instead, also a contact torque component is given. 

In order to include in the \gls{IP} the wrench polytopes $\mathcal{A}_i$ of the $\nth{i}$ limb in contact with the environment, we reformulate ~\eqref{eq:staticForcepolytope} as follows:
	\begin{equation}
	\mathcal{A}_i = \Big\{ \vc{f}_i \in \Rnum^m \quad | \quad \vc{J}(\vc{q}_i)^T \vc{f}_i = \vc{g}(\vc{q}_i) - \vc{\tau}_i, \quad \vc{\underline{\tau}}_i \leq \vc{\tau}_i \leq \vc{\overline{\tau}}_i \Big\}
	\label{eq:staticForcePolygonCompact}
	\end{equation}
	
{Considering that the joint-space torque variable $\vc{\tau}_i$ is limited by its minimum and maximum values $\vc{\underline{\tau}}_i$ and $\vc{\overline{\tau}}_i$, we can then further re-write ~\eqref{eq:staticForcePolygonCompact} to explicitly highlight this dependence}:
	\begin{equation}
	\mathcal{A}_i = \Big\{ \vc{f}_i \in \Rnum^m \quad | \quad \underbrace{\vc{g}(\vc{q}_i) + \vc{\underline{\tau}}_i}_{ \vc{\underbar{d}}_i \in \Rnum^{n_l}} \leq \vc{J}^T_i \vc{f}_i \leq \underbrace{\vc{g}(\vc{q}_i)+\vc{\overline{\tau}}_i}_{ \vc{\bar{d}}_i \in \Rnum^{n_l}} \Big\}
	\label{eq:staticForcePolyInequality}
	\end{equation}
Equation~\eqref{eq:staticForcePolyInequality} thus represents a compact notation for the static wrench polytope initially defined in ~\eqref{eq:staticForcepolytope}. We will now exploit the notation in~\eqref{eq:staticForcePolyInequality} to introduce the matrix {$\vc{G} \in \Rnum^{d \times (m n_c)}$ and the vector $\vc{d}$, result of the concatenation of all the matrices $\vc{J}(\vc{q}_i)^T$ and vectors $\vc{d}_i = [\vc{\bar{d}}^T_i, \vc{\underbar{d}}^T_i]^T \in \Rnum^{2 n_l}$ of all the individual limbs in contact with the environment:
\begin{equation}
	\begin{aligned}
	\vc{G} = \text{diag}\Big( &\mat{\vc{J}(\vc{q}_1)^T\\-\vc{J}(\vc{q}_1)^T}, \dots, \mat{\vc{J}(\vc{q}_{n_c})^T \\ - \vc{J}(\vc{q}_{n_c})^T} \Big) \in \Rnum^{d \times (m n_c)}\\
	& \vc{d} = \mat{ \vc{d}^T_1 \dots  \vc{d}^T_{n_c}}^T \in \Rnum^{2 n_c n_l}
	\end{aligned}
	\label{eq:actuationInequalities}
\end{equation}}
$\vc{G}$ and $\vc{d}$ can now be used to redefine the set $\mathcal{A}$ of actuation-consistent \gls{CoM} positions and contact forces/wrenches that satisfy all the individual wrench polytopes $\mathcal{F}_i$ for $k = 1, \dots, n_c$:
\begin{equation}\label{eq:feasibleSetActuation}
	\begin{aligned}
	\mathcal{A} = \Big\{\vc{f}\in\Rnum^{m  n_c},\quad & \vc{c}_{xy} \in\Rnum^2  \quad | \quad  \vc{A}_1\vc{f} + \vc{A}_2\vc{c}_{xy} = \vc{u}, \\ & \vc{G} \vc{f} \leq \vc{d}  \Big\}
	\end{aligned}
	\end{equation}
	In analogy with ~\eqref{eq:supportRegionBretl}, we can define a new set of actuation-consistent \gls{CoM} positions called \textit{actuation region}:
	\begin{equation}\label{eq:supportRegionActuation}
	\mathcal{Y}_a = \Big\{\vc{c}_{xy} \in\Rnum^2 \quad | \quad \exists \vc{f}\in\Rnum^{m  n_c} \st (\vc{c}_{xy}, \vc{f}) \in \mathcal{A} \Big\}
	\end{equation}

	\begin{figure}
		\centering
		\begin{subfigure}{4cm}
			\includegraphics[width=1\textwidth]{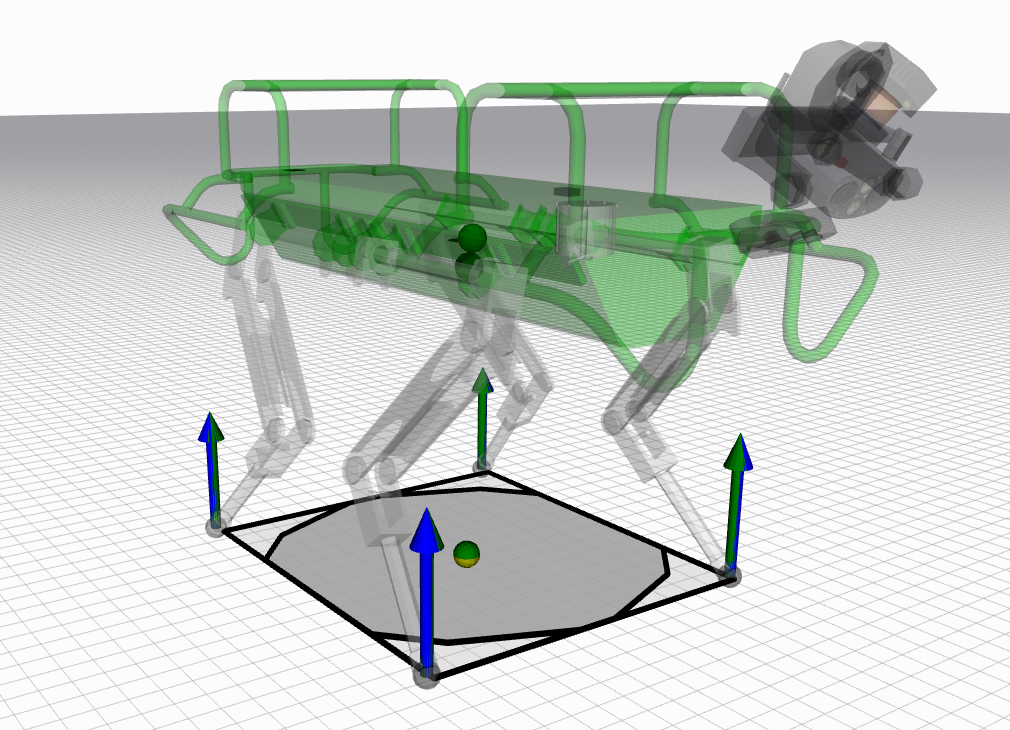}
			\caption{quadruple stance}
			\label{fig:four_stance}
		\end{subfigure}
		~
		\begin{subfigure}{4cm}
			\includegraphics[width=1\textwidth]{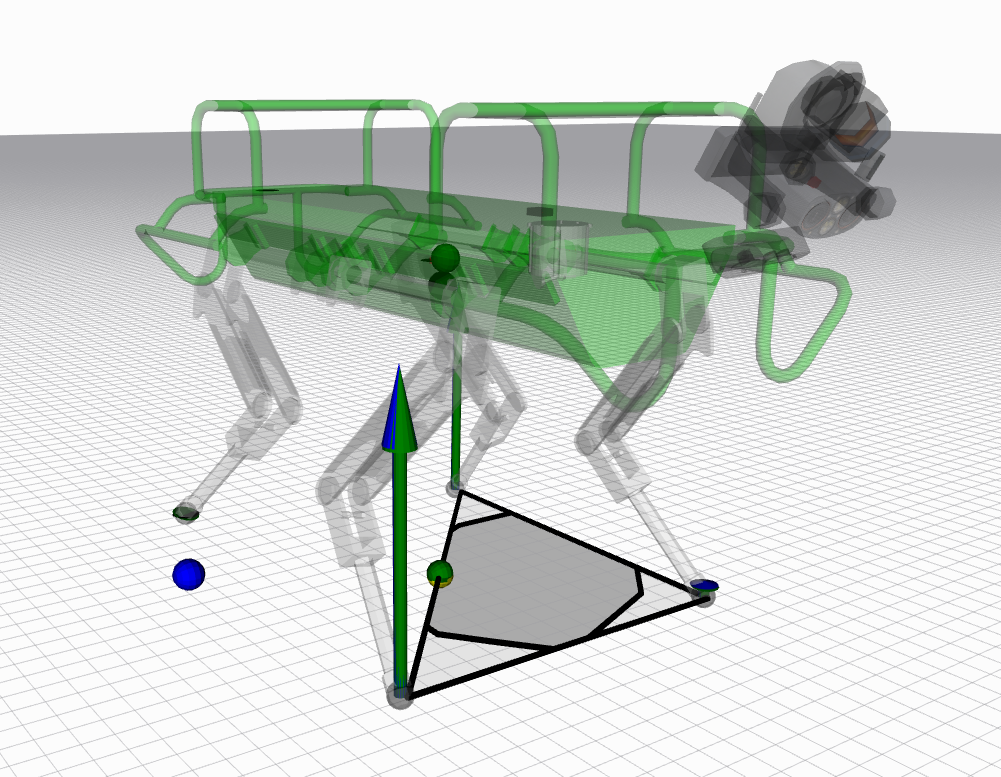}
			\caption{triple stance}
			\label{fig:triple_stance}
		\end{subfigure}
		
		\captionsetup{singlelinecheck=off}
		\caption[Friction- and actuation-consistent region]{Classical friction region (light gray) and feasible region (dark gray) in four-stance (a) and triple-stance (b) conditions.}
		\label{fig:localActuationRegion}
	\end{figure}
	As a further observation we notice that we are interested in computing the set of \gls{CoM} positions $\mathcal{Y}_{fa}$ that simultaneously satisfies both the friction and the actuation constraints (see Fig. \ref{fig:localActuationRegion}). This can be obtained by considering the intersection of $\mathcal{C}$ and $\mathcal{A}$:
	\begin{equation}\label{eq:feasibleSetFrictionAndActuation}
	\begin{aligned}
	\mathcal{C} \cap \mathcal{A} = \Big\{\vc{f}\in\Rnum^{m n_c}, &\quad \vc{c}_{xy} \in\Rnum^2 \quad | \quad \vc{A}_1\vc{f} + \vc{A}_2\vc{c}_{xy} = \vc{u}, \\
	 & \vc{B}\vc{f} \leq \vc{0}, \quad \vc{G} \vc{f} \leq \vc{d}  \Big\}
	\end{aligned}
	\end{equation}
	Based on ~\eqref{eq:feasibleSetFrictionAndActuation}, the friction- and actuation-consistent region $\mathcal{Y}_{fa}$, called \textit{feasible region}, can be defined as:
	\begin{equation}\label{eq:supportRegionActuationAndFriction}
	\mathcal{Y}_{fa} = \Big\{\vc{c}_{xy} \in\Rnum^2 \quad | \quad \exists \vc{f}\in\Rnum^{m n_c} \st (\vc{c}_{xy}, \vc{f}) \in \mathcal{C} \cap \mathcal{A} \Big\}
	\end{equation}
	In analogy with Alg. \ref{alg:iterativeProjectionBretl}, Alg. \ref{alg:iterativeProjectionWithActuation} explains how $\mathcal{Y}_{fa}$ can be computed efficiently with an iterative projection.
	
Simultaneously imposing the inequality constraints (III.b) and (III.c) in Alg. \ref{alg:iterativeProjectionWithActuation} corresponds to performing an intersection of the friction cone $\mathcal{C}_i$ with the polytopes $\mathcal{A}_i$ of the corresponding contact point. This yields the set of all the contact forces that simultaneously respect both the friction cone constraints and the joint actuation limits of the $\nth{i}$ limb (see for example Fig. \ref{fig:actuationAndFriction}). Alg. \ref{alg:iterativeProjectionWithActuation}, in practice, is equivalent to Alg. \ref{alg:iterativeProjectionBretl} with the only difference being the constraint (III.c) relative to the actuation limits.
	
The \textit{actuation region} $\mathcal{Y}_{a}$ (that only considers actuation constraints and no friction constraints) can be obtained by simply removing the constraint (III.b) from the \gls{lp} that is solved at the step III of Alg. \ref{alg:iterativeProjectionWithActuation}. {Intermediate cases exist where some end-effector present unilateral contacts and other limbs present instead bilateral contacts. This is the case, for example, when a robot is climbing a ladder pushing with its feet and pulling with his hands. Such conditions can be captured by the presented \gls{IP} modification by enforcing only the wrench polytope constraints on the bilateral contact points and by enforcing both friction pyramids and wrench polytopes on unilateral contacts. }
	
The  wrench polytope $\mathcal{A}_i$, unlike the friction cones $\mathcal{C}_i$, is a {pose-dependent} quantity and, as a consequence, its vertices will change whenever the robot changes its configuration. The feasible region $\mathcal{Y}_{fa}$ can thus be considered to be accurate only in a \textit{neighborhood} of the current robot configuration. The distance between the current \gls{CoM} projection $\vc{c}_{xy}$ and the edges of $\mathcal{Y}_{fa}$ can be considered as a combined measure of the {instantaneous} robustness of the robot's state with respect to the contacts' stability and joint-space torque limits. This distance (in [m]) can also be seen as a robustness measure against possible external loads being added on top of the robot that may move the robot's \gls{CoM} even when its configuration does not change.
	
%	Friction cones (and the friction region $\mathcal{Y}_f$) only depend on the contact configuration and can thus be recomputed only at stance change. This is a convenient property to embed a notion of \emph{contact stability}~\cite{Trinkle1997} in motion planning. Meanwhile, 
{Wrench polytopes (and thus the feasible region $\mathcal{Y}_{fa}$), because of their \textit{local validity}, must be recomputed at every configuration change which makes the motion planning formulation harder compared to cases where only the friction cones are considered. However, this local validity is also the key element of the wrench polytopes that, if properly exploited, can provide an insightful view on the relationship between robot configuration and maximal admissible force at the end-effectors.}

\subsection{2D Feasible Regions vs. 6D Feasible Polytopes}\label{sec:2Dversus6D}
To achieve a better understanding of feasible regions, it is useful to underline the parallel that exists between them and their 6D counterparts (see Tab. \ref{tab:4}). In particular, the friction region $\mathcal{Y}_f$ can be seen as a particular case of the {\acrfull{cwc}} criterion with only gravity acting on the \gls{CoM} of the robot. In the same way, also the actuation region $\mathcal{Y}_a$ can be seen as a static case of the {\acrfull{AWP}} and the feasible region $\mathcal{Y}_{fa}$ can be seen as a static case of the {\acrfull{FWP}} \cite{Orsolino2018}.
	
It is possible to show that, for example, a 2D region can be obtained from the relative 6D polytope (\eg \gls{AWP} or \gls{FWP}) by slicing the latter in correspondence of the planes: $f_x = 0, f_y = 0, f_z = mg$ and $\tau_z = 0$. In this way only two \gls{DoFs} are left which correspond to the $\tau_x$ and $\tau_y$ coordinates of the wrench space. The two-dimensional region that results from this slicing procedure can then be mapped through ~\eqref{eq:staticMapping} into a set of feasible \gls{CoM} coordinates $\vc{c}_{xy}$ that corresponds to the relative region (\eg $\mathcal{Y}_a$ or $\mathcal{Y}_{fa}$). 

Computing the \gls{AWP} or the \gls{FWP}, however, can be computationally demanding because of the high dimensionality and large amount of halfspaces and vertices. This is what motivated us to propose a variant of  the IP algorithm that allows to directly map joint-torques constraints into 2D \gls{CoM} limits (without the need of constructing first the whole 6D polytopes). This results in a computation of the 2D feasible region which is at least 20 times faster in presence of three point contacts and 50 times faster in the case of four point contacts. More details about the computational time of the feasible region are provided in the next Section.

	\begin{table}[h!]
		\begin{center}
			\begin{tabular}{ | c | c | c | }
				\hline
				constraint & (static) & (dynamic) \\ type: & 2D \gls{CoM} proj. space & $6D$ \gls{CoM} wrench\\ \hline
				friction & friction/support reg. $\mathcal{Y}_f$ &  Contact Wrench \\
				& & Cone (CWC) \\
				 \hline
				joint-torques & actuation reg. $\mathcal{Y}_a$ &  Actuation Wrench \\
				& & Polytope (AWP) \\ \hline
				friction \& joint-torques & feasible reg. $\mathcal{Y}_{fa}$ &  Feasible Wrench \\
				& & Polytope (FWP) \\ \hline
			\end{tabular}
			\caption{Analogies between 2D regions and 6D polytopes.}
			\label{tab:4}
		\end{center}
	\end{table}
		
	\subsection{Computation Time}
	The usage of the \gls{IP} algorithm implies a significant speed up for the computation of the actuation region reaching average computation times in the order of milliseconds (see Fig. \ref{fig:solveTimesHistogram}) which makes it suitable for online motion planning. \vspace{.1cm}
	
	\noindent
	\begin{minipage}{\linewidth}% to keep image and caption on one page
		\makebox[\linewidth]{%        to center the image
			\includegraphics[keepaspectratio=true,scale=0.6]{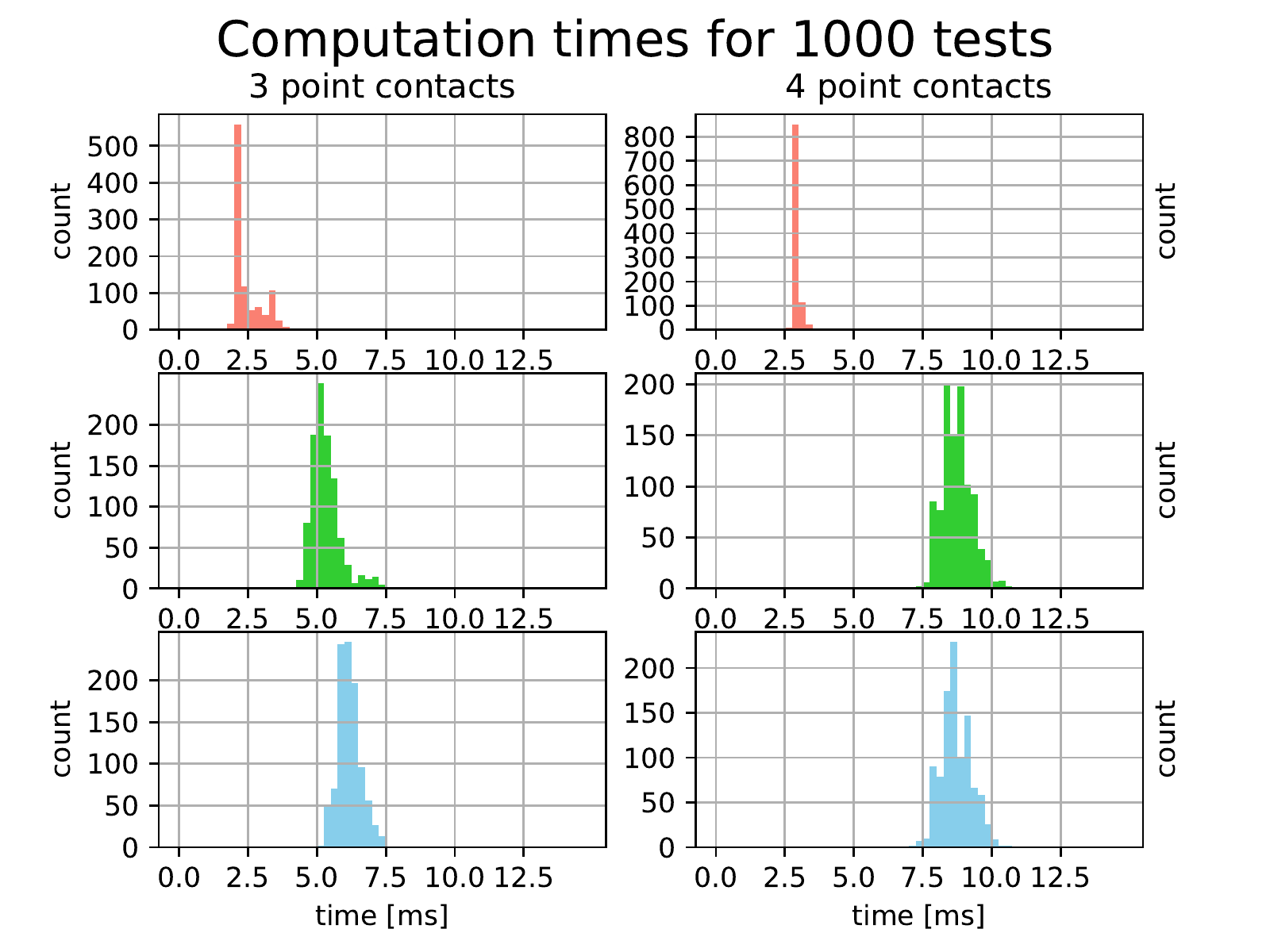}}
		\captionof{figure}[Histogram of Iterative Projection solve times]{Computation time of the IP algorithm with only friction cone constraints (red), only wrench polytope constraints (green) and both friction and actuation constraints (blue). These statistics were collected on a {8-core Intel Xeon(R) CPU E3-1505M v6 @ 3.00 GHz computer}.}
		\label{fig:solveTimesHistogram}
	\end{minipage}
\begin{figure}[b]
		\centering
		\begin{subfigure}{3.5cm}
			\includegraphics[width=1\textwidth]{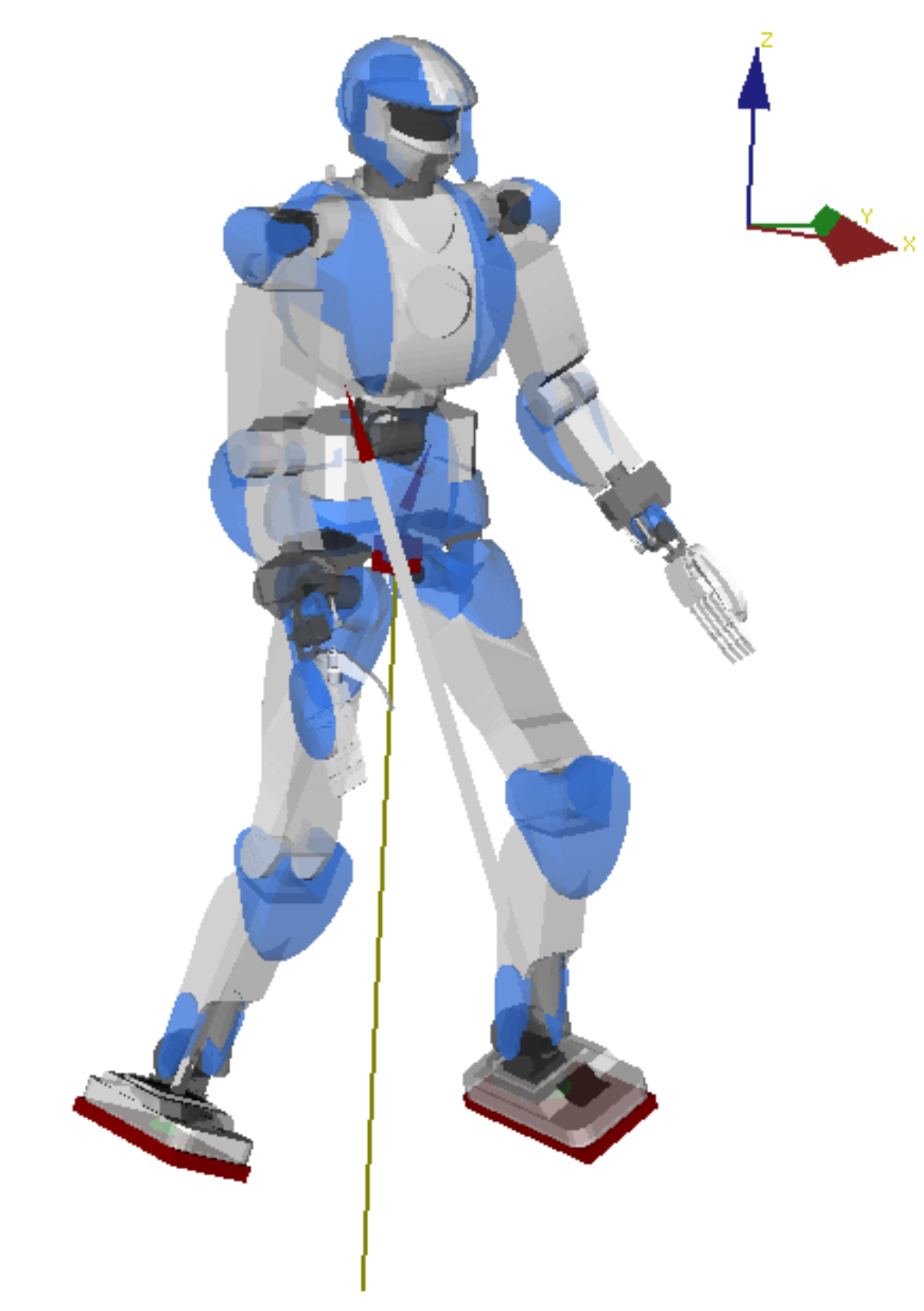}
			\caption{}
			\label{fig:HRP4_lateral}
		\end{subfigure}
		~
		\begin{subfigure}{4.5cm}
			\includegraphics[width=1\textwidth]{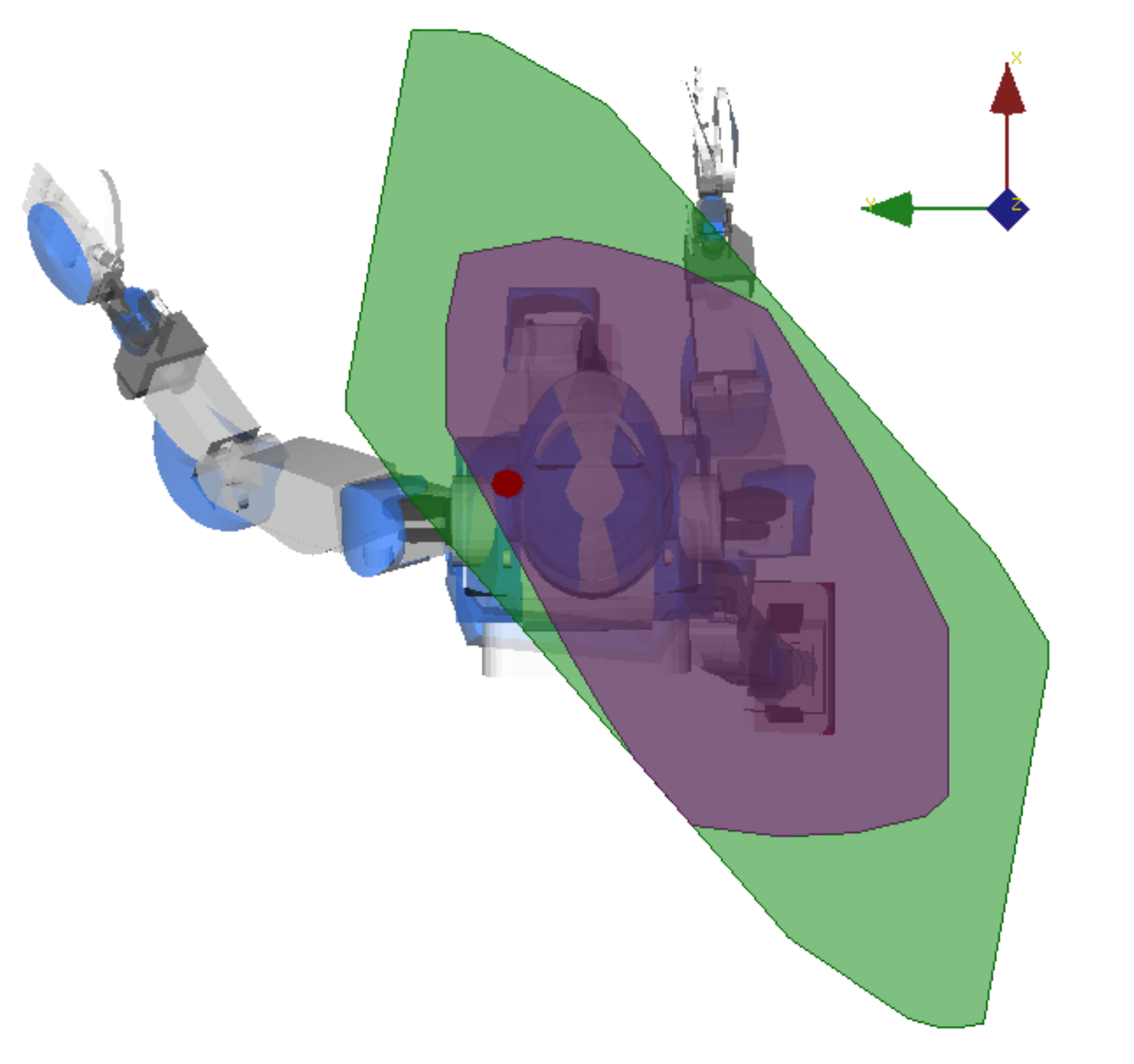}
			\caption{}
			\label{fig:HRP4_above}
		\end{subfigure}
		
		\captionsetup{singlelinecheck=off}
		\caption[Feasible region for the HRP-4 humanoid robot]{Feasible region (blue) and friction region (green) for the HRP-4 humanoid robot in a configuration with non-coplanar contacts.}
		\label{fig:humanoidRegionHRP}
	\end{figure}
	
	The solve time of the \gls{IP} algorithm depends on the number of inequality constraints embedded in it (only friction constraints, only actuation constraints, or both). The most favorable scenario is when only friction cones are considered (red in Fig. \ref{fig:solveTimesHistogram}): in the case of linearized friction cones with four facets per pyramid, the \gls{IP} will present $4 n_c$ inequalities. The least convenient scenario is instead when both friction pyramids and  wrench polytope constraints are considered (blue in Fig. \ref{fig:solveTimesHistogram}), in this case the \gls{IP} will include $(4 + 2n_l) n_c$ inequalities (assuming that all the limbs in contact with the ground have same number of \gls{DoFs} $n_l$ and that the friction cones are linearized with 4 halfspaces). In the case of the \gls{HyQ} quadruped this will result in $10$ inequalities per foot contact; in the case of a humanoid robot with 6 \gls{DoFs} per leg, instead, this will result in $16$ inequalities per foot contact. Figure \ref{fig:humanoidRegionHRP}, for example, shows the friction region $\mathcal{Y}_f$ (green) and the feasible region $\mathcal{Y}_{fa}$ in the case of the HRP-4 robot standing still in a configuration with non-coplanar contacts. 

The last row of Fig. \ref{fig:solveTimesHistogram} shows that, even in such inconvenient condition where all contacts are subject to both friction and actuation constraints, the solve time is below $10ms$ in a four-stance configuration and below $7.5ms$ in a triple-stance configuration in $99.5\%$ of the computations (blue histogram). This allows the efficient computation of the feasible region at a frequency of, at least, \SI[inter-unit-product =\ensuremath{\cdot}]{100}{\hertz} in a four stance configuration and \SI[inter-unit-product =\ensuremath{\cdot}]{133}{\hertz} in a triple stance configuration of a quadruped robot\footnote{These computation times, as much as the other performance reported in this manuscript, have been achieved on a {8-core Intel Xeon(R) CPU E3-1505M v6 @ 3.00GHz computer}.}. These frequencies could be further increased by reducing the tolerance factor of the \gls{IP} algorithm (the tolerance value we used was $10^{-6} m^2$).

\subsection{The Feasible Region under Different Loading Conditions}
\begin{figure}[h]
		\centering
		\begin{subfigure}{4.2cm}
			\includegraphics[width=1\textwidth]{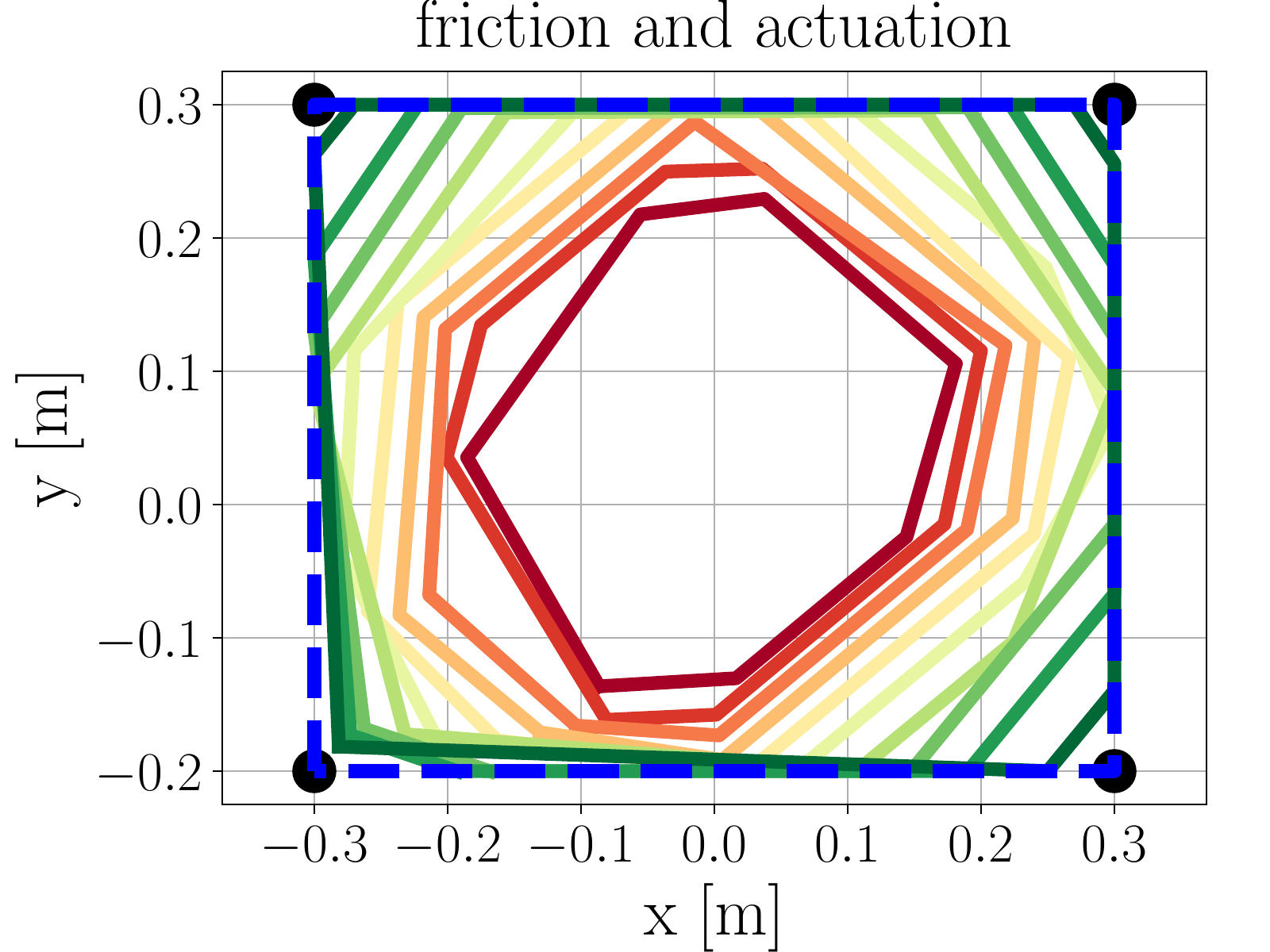}
			\caption{}
			\label{fig:feasible4}
		\end{subfigure}
		~
		\begin{subfigure}{4.2cm}
			\includegraphics[width=1\textwidth]{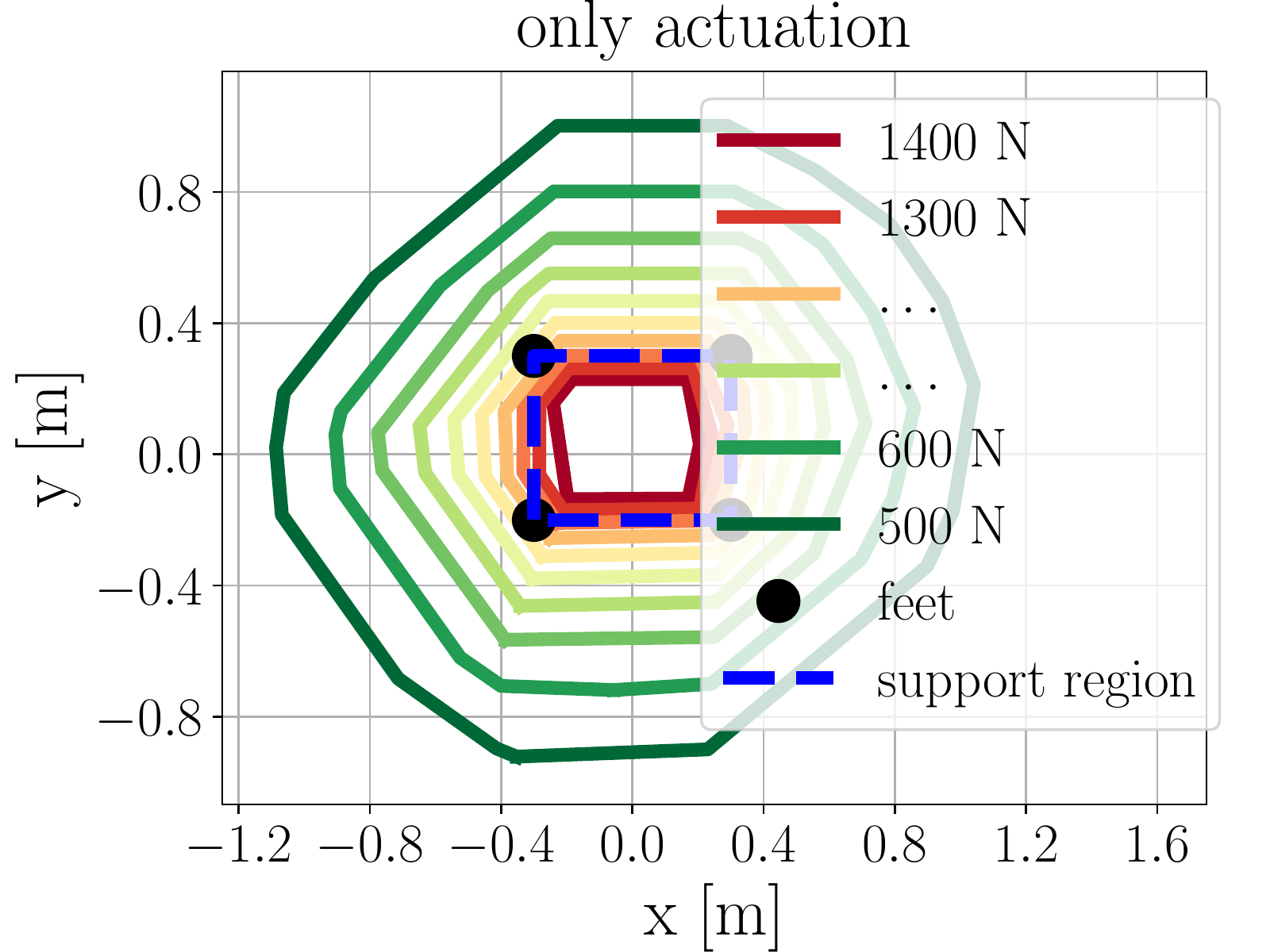}
			\caption{}
			\label{fig:actuation4}
		\end{subfigure}
		\begin{subfigure}{4.2cm}
			\includegraphics[width=1\textwidth]{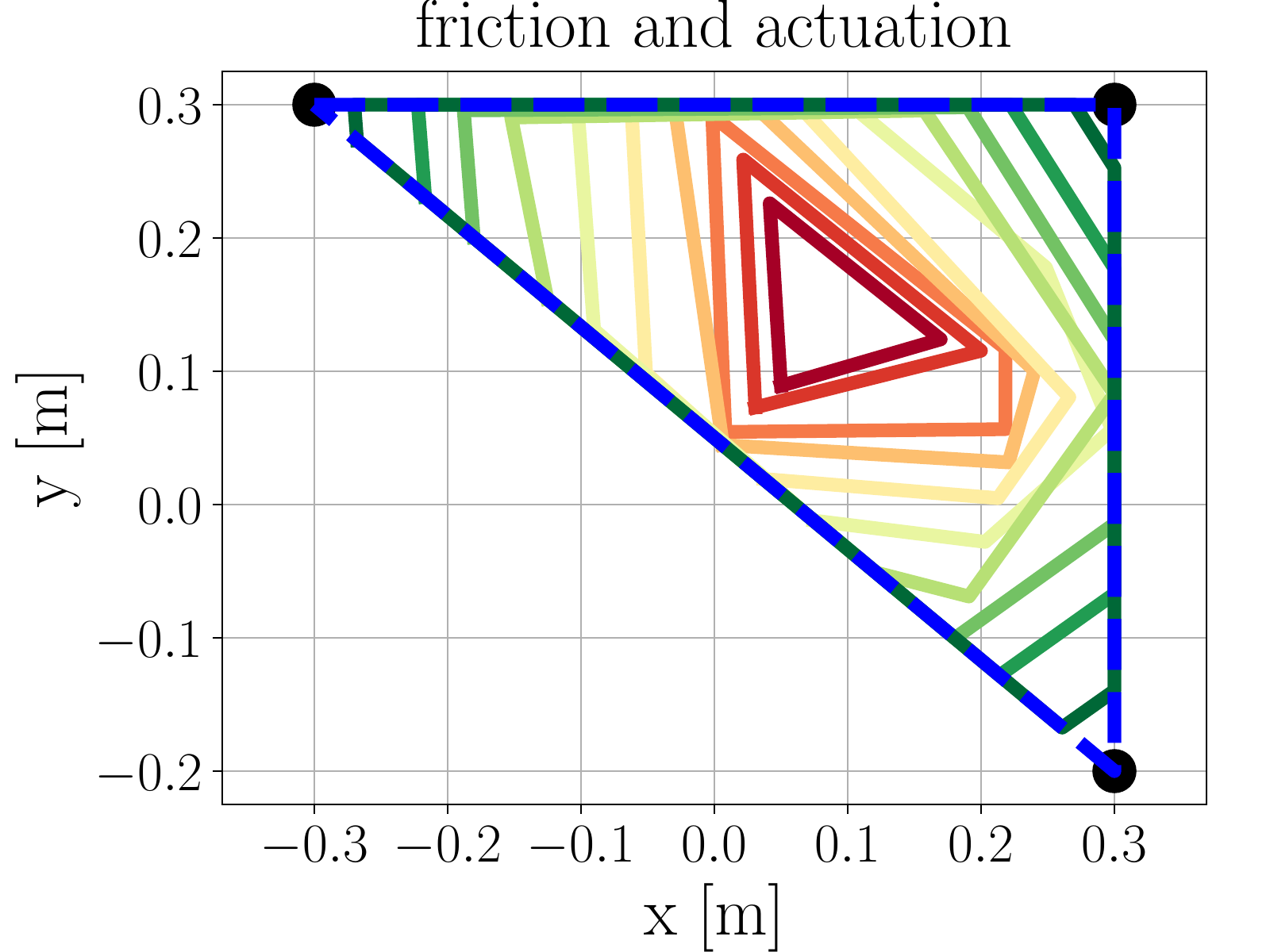}
			\caption{}
			\label{fig:feasible3}
		\end{subfigure}
		~
		\begin{subfigure}{4.2cm}
			\includegraphics[width=1\textwidth]{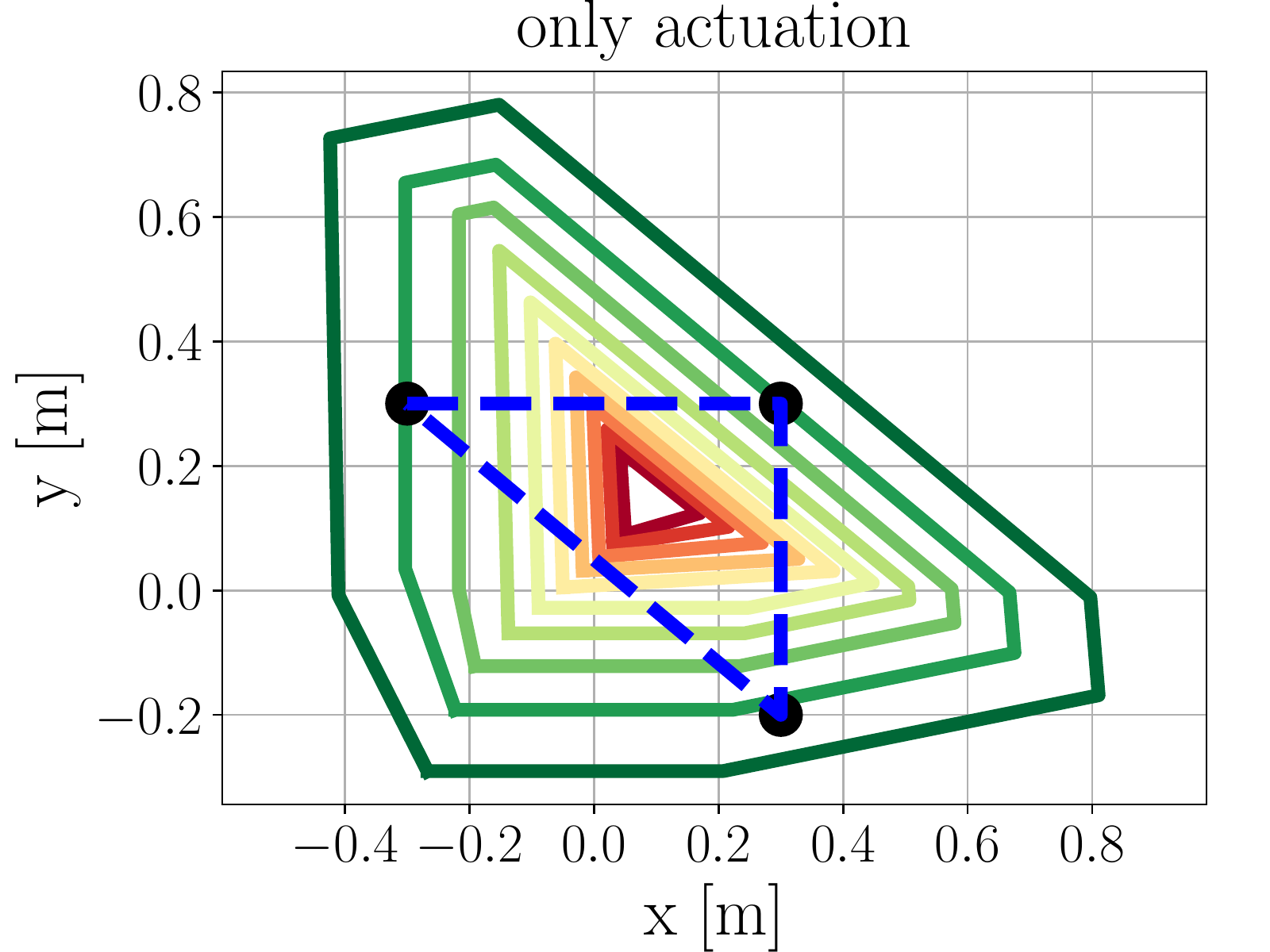}
			\caption{}
			\label{fig:actuation3}
		\end{subfigure}
		
		\captionsetup{singlelinecheck=off}
		\caption[Local actuation and feasible area examples]{{Every 2D polygon in this figure represents a feasible region computed for a different gravitational load acting on the robot's \gls{CoM}. The feasible regions are computed in four possible scenarios:
		\begin{enumerate}[(a)]
		\item 4 stance feet $\&$ friction and actuation constraints;
		\item 4 stance feet $\&$ actuation constraints (no unilaterality);
		\item 3 stance feet $\&$ friction and actuation constraints;
		\item 3 stance feet $\&$ actuation constraints (no unilaterality);
		\end{enumerate}
	The black points represent the stance feet positions of the \gls{HyQ} quadruped during a four and triple support phases; the dashed blue lines represent the feasible region obtained by consideration of friction constraints only. }
}
		\label{fig:variableMassActuationRegions}
	\end{figure}

Fig. \ref{fig:variableMassActuationRegions} reports various tests of computation of feasible 2D areas for different {gravitational} loads applied on the \gls{CoM} of the robot. {We can see that the heavier the load on the robot, the smaller the area of the corresponding feasible region.} This is analogous to the computation of feasible regions for different percentages of torque limits while keeping the load on the robot fixed. The blue dashed lines represent the classical friction region $\mathcal{Y}_f$ as defined by Bretl \etal \cite{Bretl2008}.\\
Figs. \ref{fig:feasible4} and \ref{fig:feasible3} depict the \textit{feasible regions} $\mathcal{Y}_{fa}$ for the \gls{HyQ} robot with four and three coplanar stance feet. 
	Figs. \ref{fig:actuation4} and \ref{fig:actuation3} depict the \textit{actuation regions} $\mathcal{Y}_{a}$ in the same configurations with four and three coplanar stance feet. Such actuation consistent areas $\mathcal{Y}_{a}$ alone are not directly applicable in the field of legged locomotion where robots typically make and break contacts using their feet and have therefore no possibility to grasp the terrain. Feasible regions $\mathcal{Y}_{fa}$ should be used instead since they include the friction constraints that also naturally encode the unilaterality constraint. 
	
	As visible in Figs. \ref{fig:actuation4} and \ref{fig:actuation3} the robot's \gls{CoM} might lean outside of the classical friction region $\mathcal{Y}_f$ (dashed blue line) depending on the magnitude of the load acting on it: this is a typical condition in which one of the contacts is meant to \textit{pull} the ground to maintain equilibrium.
	
	As a final consideration, comparing the figures related to the same number of stance feet (Fig. \ref{fig:actuation3} compared with \ref{fig:feasible3} and Fig. \ref{fig:actuation4} compared with \ref{fig:feasible4}) one can see that the feasible region $\mathcal{Y}_{fa}$ cannot be obtained by simple intersection of the friction region $\mathcal{Y}_{f}$ and the actuation region $\mathcal{Y}_{a}$. Although this approximation might be accurate under specific conditions, in general the intersection and projection operators do not commute \cite{Kelley1960}. Let us consider $\mathcal{C}$ to be the set of contact forces and \gls{CoM} positions $\vc{c}_{xy}$ that respect static equilibrium and friction constraints (see ~\eqref{eq:feasibleSetBretl}); let us then also consider $\mathcal{A}$ defined as the set of contact forces and horizontal \gls{CoM} positions $\vc{c}_{xy}$  that respect all static equilibrium, wrench polytopes and friction cones constraints (see ~\eqref{eq:staticForcePolyInequality}). The \textit{friction region}~\cite{Bretl2008} can then be defined compactly as:
	\begin{equation}
	\mathcal{Y}_f = IP(\mathcal{C}),
	\end{equation}
	the \textit{actuation region} as:
	\begin{equation}
	\mathcal{Y}_a = IP(\mathcal{A})
	\end{equation}
	and the (actuation- and friction-consistent) \textit{feasible region} as:
	\begin{equation}
	\mathcal{Y}_{fa} = IP(\mathcal{C} \cap \mathcal{A})
	\end{equation}
	where $IP$ is the Iterative Projection operator. The projection and intersection are non-commutative operators and, in particular, the following inclusion always holds:
	\begin{equation}
	\mathcal{Y}_{fa} \subseteq \mathcal{Y}_{f} \cap \mathcal{Y}_{a}
	\label{eq:setsInclusion}
	\end{equation}
	$\mathcal{Y}_{fa}$ is therefore more conservative than the intersection of $\mathcal{Y}_{f}$ and $\mathcal{Y}_{a}$. Intuitively, ~\eqref{eq:setsInclusion} might be explained by considering that there may exist \gls{CoM} positions that, at same the time provide feasible wrench solutions if the friction cones or wrench polytope constraints are considered \textit{individually} but they provide unfeasible wrench solutions if these constraints are considered \textit{simultaneously}. 
	As a consequence, the feasible region $\mathcal{Y}_{fa}$ has to be a subset of the intersection between $\mathcal{Y}_{f}$ and $\mathcal{Y}_{a}$.
\section{Center of Mass and Foothold Planning}\label{sec:CoMandFeetPlanning}
In this Section we employ the concept of feasible region $\mathcal{Y}_{fa}$ introduced in Section \ref{sec:feasibleRegion} for the sample-based optimization of feasible footholds and \gls{CoM} trajectories. {The all locomotion scheme proposed in this Section is based on the assumption that a robot is assumed to be statically stable and, simultaneously, a set of admissible joint-torques exists iff:
\begin{equation}\label{eq:stabilityCriterion}
	\vc{c}_{xy} \in \mathcal{Y}_{fa}
\end{equation}
where $\vc{c}_{xy} = \vc{P} \vc{c}$ is the \gls{CoM} projection on the horizontal plane.
\textit{Robustness} against possible modelling errors or external disturbances can then be enforced by considering 
the minimum distance $r$ between $\vc{c}_{xy}$} and the edges of the feasible region. This can be found by solving the following LP:
\begin{argmaxi}|l|
	  {r}{ \vc{a}^T_i \vc{c}_{xy} + ||\vc{a}_i||_2 r \leq b_i, \quad i=0,\dots,N_h}{}{}
	  \label{eq:feasibilityMargin}
\end{argmaxi}
where $N_h$ is the number of edges of $\mathcal{Y}_{fa}$, $\vc{a}_i \in \Rnum^2$ is the normal to the $\nth{i}$ edge and $b_i \in \Rnum$ is the known term. 
$r$ is thus the radius of the largest ball centered in $\vc{c}_{xy}$ and inscribed inside $\mathcal{Y}_{fa}$.
and it can also be seen as a static instantaneous measure 
(\ie a margin) of how far the robot is from slipping or from hitting one of joint-torque limits (actuation limits). 

\subsection{CoM planning strategy}\label{sec:CoMplanning}
As we only deal with \gls{CoM} planning in this Section, we will assume the gait sequence, 
phase timings and step locations to be predefined. 
Since the feasible region, at the actual state, is restricted by the quasi-static assumption
(an extension to the dynamic case with non-negligible \gls{CoM} horizontal acceleration is part of future works)
a quasi-static gait is a good template 
to test its applicability.  

As the main hardware platform for our experiments is  the quadruped robot \gls{HyQ}, 
we will consider here a static quadrupedal gait called \textit{crawl} \cite{Focchi2018}. 
The crawl is typically divided in two main phases called \textit{swing phase} and \textit{move-base phase}. During the swing phase, the robot does not 
move its trunk and only one foot at the time is allowed to lift-off from the ground and move to a new foothold while all the other three feet have to be in stance. During the move-base phase, instead, all four feet are in stance and the robot moves its trunk to a target location and orientation.

The most critical phase of this static gait, in terms of stability and margin with respect to the joint-torque limits, is this triple stance phase (\ie the swing phase) because the robot's weight 
must be distributed only on three legs.
The \gls{CoM} is meant to move only during the four-stance phase, to enter  the future support region which is opposite to the next swing leg. 
Therefore, after each touchdown, we re-plan a polynomial trajectory that links the actual \gls{CoM} position with a new target  
inside the future support region. This enables us to completely unload the swing leg before 
liftoff and naturally distribute the weight onto the other three stance legs.
In our previous work \cite{Focchi2018} we computed this target heuristically without any awareness of joint-torque limits. 
Specifically, we were computing the target point at a \textit{hand-tuned} distance from the main diagonal of the support triangle, 
in order to \textit{sufficiently} load the off-diagonal leg. 
However, this can be inaccurate in complex terrains, because:
\begin{enumerate}
	\item the friction region $\mathcal{Y}_f$ coincides with the feasible region $\mathcal{Y}_{fa}$ only when every individual limb of the robot is able to carry the total body weight of the robot;
	\item an increased load on the robot or an inconvenient robot configuration can further restrict the feasible region $\mathcal{Y}_{fa}$ making it considerably smaller than the friction region. 
\end{enumerate}
Therefore, the heuristic target, since it is not formally taking these aspects into account, might fail in situations that are more demanding due to a complex terrain geometry.
{Conversely, using the feasible region allows us to select a target position for the robot's \gls{CoM} that results in an admissible and statically stable configuration, in the case of:
1) a generic terrain shape (i.e. non coplanar feet, each one with different normal at the contact)
2) different loading conditions. }

{We plan the \gls{CoM} target to be inside a \emph{scaled} feasible region by a predefined scaling factor $s$. This allows us to increase the \emph{robustness} of our strategy against possible external disturbances and modelling inaccuracies that may arise from the static locomotion assumption that we took in~\eref{eq:staticForcepolytope}\footnote{{This has a particular effect on the execution of dynamic motions on electrically-actuated robots whose torque limits (and thus their wrench polytopes) depend on the joint velocity.}}.}

At the touch-down instant, we compute the feasible region $\mathcal{Y}_{fa}$, considering as inputs 
the position of the three stance feet of the future support triangle (the feet sequence is predefined) and the corresponding normals $\vc{n}_i$ at the expected contact points. To evaluate the Jacobians (necessary to map the actuation constraints into a set of admissible contact forces), we also provide the future \gls{CoM} position predicted by the heuristics. If the projection of the actual \gls{CoM} $\vc{c}_{xy} = \vc{P} \vc{c}$ is inside $\mathcal{Y}_{fa}$, we then set the target \gls{CoM} equal to the actual \gls{CoM} $\vc{c} \in \Rnum^3$. If it is, instead, outside the region $\mathcal{Y}_{fa}$, we set the target \gls{CoM} equal to the point $\vc{x}^*$ on the boundary of the 
region (or of the scaled region if we want to provide a certain degree of robustness)
that is closest to $\vc{c}_{xy}$. This allows us to minimize unwanted lateral/backward motions. 
To obtain the point $\vc{x}^*$ we solve the following QP program:
\begin{align}
\vc{x}^* & = \quad \argmin_{\vc{x} \in \Rnum^2}{\Vert \vc{x} - \vc{P} \vc{c}\Vert^2} \\
\text{subject to:}& \quad \vc{A} \vc{x} \leq \vc{b}
\end{align}

where we minimize the Euclidean distance between a generic inner point $\vc{x}$ and the actual \gls{CoM} projection $\vc{c}_{xy}$. 
$\vc{A}$ and $\vc{b}$ matrix represent the half-space description of the polygon $\mathcal{Y}_{fa}$.\\
The \gls{CoM} target is depicted as a yellow cube in Fig. \ref{fig:footholdPlanning} 
while the blue cube represents the heuristic target.
In the same picture we show an image of the feasible region $\mathcal{Y}_{fa}$ (light gray) and the scaled feasible region (dark gray) scaled by a factor of $s = 0.8$. The dashed triangle represents the friction region $\mathcal{Y}_f$\footnote{Note that just scaling the value of joint-torque limits (instead of the vertices of the feasible region) might not results in a conservative region. This is because 
some boundaries of the resulting feasible region could be determined by the friction region itself, thus reducing the joint-torque limits would not result in an increase of robustness with respect to those boundaries. For this reason, it is advisable to scale directly the vertices of the feasible region rather than joint-torque limits used to compute the feasible region.}.
The scaling procedure can be defined as an affine 
transformation that preserves straight lines and parallelism relationships among the 
edges of the feasible region. 
%For the transformation to preserve the angles, it should be performed 
%as a scaling of the vertices of the polygon with respect to the Chebyshev center (\ie the center 
%of the largest ball inscribed in the feasible region). 
The scaling can be done with respect to the Chebyshev center (\ie the center 
of the largest ball inscribed in the feasible region) or with respect to the centroid. 
The former is more computationally expensive because it requires the solution of an \gls{lp}; the latter is faster to compute because it can be found 
analytically as the average of all the vertices. The centroid $v_c$  can be considered as a 
good approximation of the Chebychev center whenever the feasible region presents good symmetry properties{\footnote{{Whenever the feasible region is not symmetric, however, the centroid might considerably differ from the Chebychev 
center thus resulting in a value of the robustness margin $r$ lower than desired.}}.
In the case that $v_c$ is used, the vertices $\hat{v} $ of the scaled region $\hat{\mathcal{Y}}_{fa}$ can be computed by scaling the  vertices $v$ of $\mathcal{Y}_{fa}$ as: $\hat{v} = s(v - v_c) + v_c $
where $s \in (0.0,1.0]$ is the scaling factor. }
%
%
%\begin{figure}
%	\centering
%	~ %add desired spacing between images, e. g. ~, \quad, \qquad, \hfill etc. 
%	%(or a blank line to force the subfigure onto a new line)
%		\includegraphics[width=0.4\textwidth]{figs/actuation_region/scaled_feasible_region}
%		\caption{Classical friction region (dashed lines), feasible region (light gray) and scaled feasible region (dark gray) with a scaling factor of $0.8$.}
%		\label{fig:scalingFeasibleRegion}
%\end{figure}
	
%The heuristic strategy that we propose is that during the triple stance phases 
%(those which are most critically for stability and feasibility compared to when the robot has four feet on the ground) the 
%\gls{CoM} projection $\vc{c}_{xy}$ is planned to lie on the edge of a conservative 
%local feasibility region $\hat{\mathcal{Y}}_{fa}$. The region $\hat{\mathcal{Y}}_{fa}$ 
%is obtained by scaling the local feasible region of a predefined percentage 
%(typically $0.9$ of scaling factor). 

%Moreover, might not explicitly enforce 
%the joint torque constraints
%\footnote{The whole-body controller might be equally 
%minimizing all the joint torques rather than explicitly constraining them inside 
%the torque limits or the employed weight matrix might penalize the torque on a 
%specific predefined joint (e.g. the knee torque) that is not the one being closest 
%to the torque limits violation for the current configuration.} \cite{focchi2016}.\\
		
\subsection{Foothold Planning}
\label{sec:footholdPlanning}
	
The foothold planning strategy that we present in this Section represents a 
sample-based strategy to improve the navigation capabilities of the \gls{HyQ} quadruped 
robot on rough terrains. Our strategy employs the height map provided by the perception 
module and seeks among the terrain samples the foot location that maximizes 
the area of the corresponding feasible region $\mathcal{Y}_{fa}$.

We exploit the computational efficiency of the \gls{IP} algorithm as in Alg. \ref{alg:iterativeProjectionWithActuation} 
in order to plan foothold locations that ensure the robot's stability and actuation consistency while traversing rough terrains. 
As in the previous Section, we assume here a static crawl gait with predefined {durations of the stance and swing phases.}
The idea is to find, at each lift-off, the most suitable foothold to maximize 
the area of the feasible region for the next swing leg. 
Our strategy consists in sampling a set of $p$ candidate footholds around the \textit{default}  target foothold 
(from heuristics) located along the direction of motion. We then evaluate
the height map of the terrain in those sampled points
(\ie correcting the corresponding $z$ coordinate and swing orientation to adapt 
to the perceived terrain surface
\footnote{To avoid corrections in unwanted directions, we define 
the sampling direction along the direction of the predicted step, 
(i.e. in consistency with the desired velocity).}) \cite{camurri17ral, nobili_camurri2017rss}.   
The default step location is simply a function of the user-defined desired linear and angular velocities 
of the robot and it neither considers the external map of the surrounding environment,
nor the stability and actuation consistency requirements \cite{Focchi2018}. 
Fig. \ref{fig:footholdPlanning} shows a foothold planning simulation 
in which eight different candidate footholds (red spheres) are considered.
As additional feature,  we discard the footholds that:
1) are close to the edge, 2) would result in a shin collision, 3) are out of the leg's workspace. 

In the simulation shown one out of 8 is discarded because it was too close to the edge of the pallet. 
The next step consists in computing the feasible regions $\mathcal{Y}_{fa}^i$ 
for the $p=8$ considered foot locations ($i = 1, \dots, p$) keeping fixed the set of feet 
that will be in stance during the following swing phase.
%For this heuristics  to yield reliable results the local feasible regions must also 
Since the feasible region depends on the robot configuration, we consider
the future position of the \gls{CoM} (computed through the heuristics) for the next triple stance phase and obtain the future joints configuration through inverse kinematics. This joint configuration is then used to update the Jacobians needed for the computation of the candidate feasible region $\mathcal{Y}_{fa}^i$.
The foothold planner then selects, among the reduced set of  
admissible footholds, the one that maximizes the area of the corresponding feasible region\footnote{Another approach could consist in maximizing the residual radius (i.e. radius of the largest circumference inscribed in the region), however, we noticed that often multiple candidate footholds may return the same residual radius but different areas. This is the case any time that the \gls{CoM} projection is closer to a friction-limited edge rather than an actuation-limited edge of the feasible region.}.

In the baseline walking on flat terrain, when joint-torques are far away from their limits, the default foothold is selected. Conversely, on more complex terrain, 
when the robot is far from a default configuration (\eg when one leg is much more retracted than the other legs),
the scaled version $\hat{\mathcal{Y}}_{fa}$ (described in the previous Section) can take on a small area 
(see Fig. \ref{fig:footholdPlanning}). 
In this case the default step will be \textit{corrected} (yellow ball in Fig. \ref{fig:footholdPlanning}) 
in order to enlarge this area 
and, as a consequence, to increase the robustness to model uncertainties and tracking errors. 
The default target is not visible because, being computed on a planar estimation 
of the terrain \cite{Focchi2018}, it turns out to be inside the {terrain}.
%
%	We then exploit the knowledge of the (decoupled) \gls{CoM} planning policy
%	to plan foothold locations that are coherent with the future reference 

	\noindent%
	\begin{minipage}{\linewidth}% to keep image and caption on one page
\makebox[\linewidth]{%        to center the image
\includegraphics[width=1.0\columnwidth]{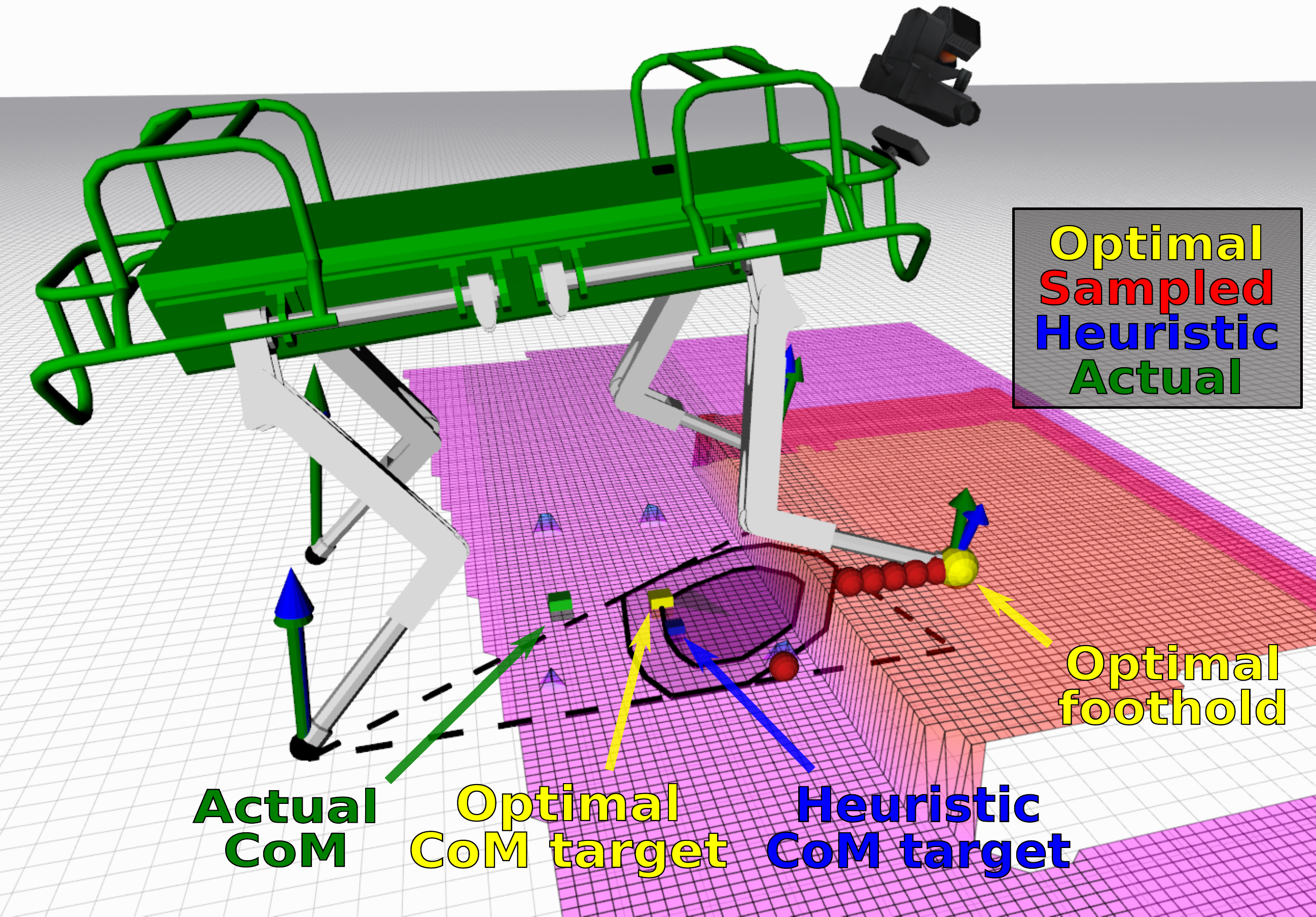}}
\captionof{figure}{{We show the classical friction region (dashed lines), the feasible region (light gray) 
	and the scaled feasible region (dark gray) with a scaling factor $s = 0.65$. The cubes represent \gls{CoM} projections on the ground of different meaning: the heuristic \gls{CoM} target (blue), the \gls{CoM} target as computed using the feasible regions (yellow) and the actual \gls{CoM} (green). The spheres represent candidate footholds (red) and the optimal foothold (yellow) selected by the sampling-based planner. We can see that the optimal \gls{CoM} target (yellow cube) lies on the edge of the scaled feasible region (dark gray) computed using the optimal foothold (yellow sphere).}}
\label{fig:footholdPlanning}
	\end{minipage}

\section{Simulation and Experimental Results}\label{sec:results}
	
The improvement of a planning strategy based on the feasible regions with respect to our previous heuristic strategy can be demonstrated by either increasing the load acting on the robot during a standard walk on a flat terrain or by addressing challenging terrains. Both scenarios, and any combination of them, bring indeed the robot closer to its actuation limits. 

As a first result we report the validation of the feasibility margin defined as the distance between the \gls{CoM} projection and the edges of the feasible region. We then report some simulation and experimental data of the \gls{CoM} and foothold strategy that we described above in Sec. \ref{sec:CoMplanning}.
The results of this strategy can be seen in the accompanying 
video\footnote{\href{https://youtu.be/9pvWO2Qmo9k}{\texttt{https://youtu.be/9pvWO2Qmo9k}}}.

\subsection{Validation of the Feasibility Margin}
Figure \ref{fig:increasingLoadSimuluation} represents the data collected in a simulation where we applied on the \gls{CoM} of the HyQ robot a vertical increasing force from \SI[inter-unit-product =\ensuremath{\cdot}]{0}{\newton} up to \SI[inter-unit-product =\ensuremath{\cdot}]{-600}{\newton} (upper plot). {This force represents a possible external payload applied on the robot and, by increasing its amplitude, we are interested in understanding how the feasible region $\mathcal{Y}_{fa}$ adapts to it.} 

{In the lower plot of Fig.  \ref{fig:increasingLoadSimuluation} we can see that, as a consequence of this increasing external load, the feasible region gradually shrinks} with a consequent reduction of the feasibility margin $r$ from \SI[inter-unit-product =\ensuremath{\cdot}]{0.24}{\meter} to about \SI[inter-unit-product =\ensuremath{\cdot}]{0.06}{\meter}.
Recall that the feasibility margin $r$ is defined as the minimum distance between the \gls{CoM} projection $\vc{c}_{xy}$ and the edges of the feasible region $\mathcal{Y}_{fa}$ (as in ~\eqref{eq:feasibilityMargin}).

For this validation we also introduce the joint-torque limits violation flag $\beta$ whose definition is the following:
\begin{equation}
\beta = \Big\{ \begin{tabular}{ll}
 0 & \text{if} \quad $\tau_i \in [\tau^{max}_i, \tau^{min}_i], \quad \forall i = 0, \dots, n$ \\
 1 & \text{otherwise} \\
\end{tabular}
\end{equation}
\noindent%
\begin{minipage}{\linewidth}% to keep image and caption on one page
\makebox[\linewidth]{%        to center the image
\includegraphics[keepaspectratio=true,scale=0.075]{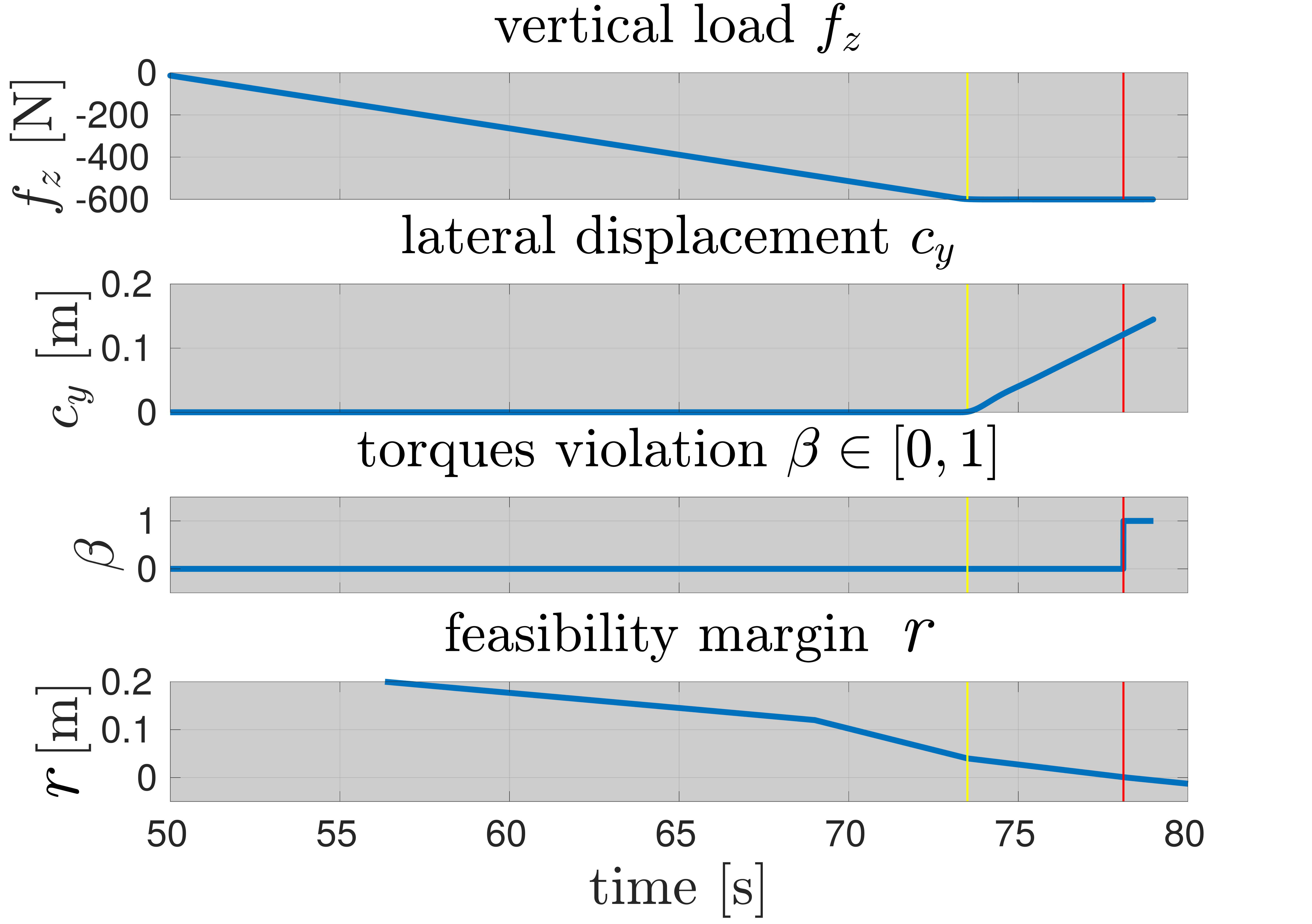}}
\captionof{figure}[]{Validation of the distance between the \gls{CoM} projection and the edges of the feasible region $\mathcal{Y}_{fa}$:  one of joint-torque limits is hit (\ie $\beta = 1$) approximately at the same time when the feasibility margin $r$ becomes negative (lower plot).}
	\label{fig:increasingLoadSimuluation}
	\vspace{+0.3cm}
\end{minipage}

\noindent Note that a negative $r$  means that the \gls{CoM} projection $\vc{c}_{xy}$ lies outside the edges of the feasible region $\mathcal{Y}_{fa}$.
After second $74$ (yellow vertical line) the external load is fixed to \SI[inter-unit-product =\ensuremath{\cdot}]{-600}{\newton} and the robot starts displacing laterally with an increasing $c_y$ coordinate. The second plot from above shows that, when the robot has moved laterally of about \SI[inter-unit-product =\ensuremath{\cdot}]{0.12}{\meter} (red vertical line), the feasibility margin $r$ becomes zero and, approximately at the same time, the torque limits violation flag $\beta$ becomes one, meaning that one of joint-torque limits of the robot has been reached (second plot from {the bottom}). 

\begin{multicols}{2}
\end{multicols}
\begin{figure*}[!]
	\includegraphics[width=\textwidth,width=18cm]{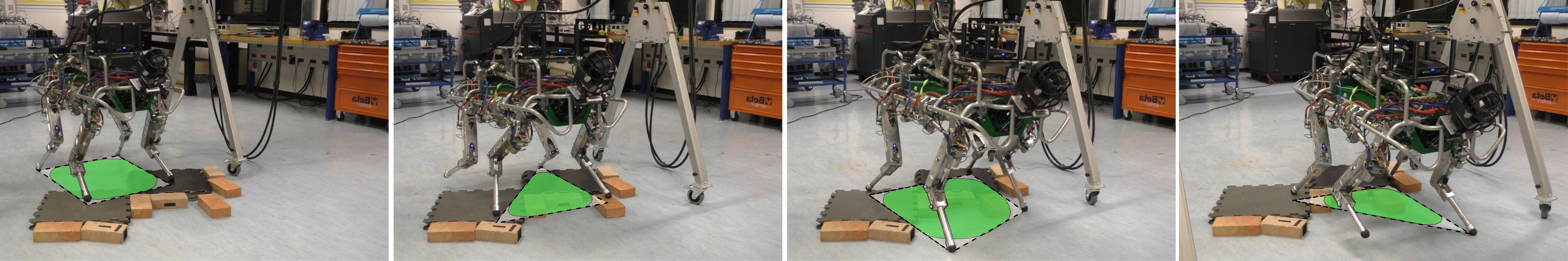}
	\caption{{Snapshots of \gls{HyQ} traversing a course of mild roughness using the map-based \gls{CoM} and foothold selection strategy presented in this manuscript. The 2D polygons represent the classical support region, or friction region, (gray area with dashed lines) and the feasible region (green area with thick lines). The rougher the terrain, the higher the difference between the support region and the feasible region.}}
	\label{fig:timelapse}
\end{figure*}
\begin{multicols}{2}
\end{multicols} 

\subsection{Walk in Presence of Rough Terrain and External Load}
The next simulation result that we report in this Section is a walk over a \SI[inter-unit-product =\ensuremath{\cdot}]{0.22}{\meter} high pallet, where the HyQ robot only lifts two lateral legs on the pallet while the two other legs always remain on the flat ground. The considerable height of the pallet and the asymmetry of the terrain force the robot to take on complex configurations to step up and down the obstacle and, even if no further external load is applied, the robot might easily reach its joint-torque limits. In this scenario we compare the behavior of two different strategies:
\begin{enumerate}
\item \textit{friction-region based walk:} this motion planning approach combines the foothold selection strategy explained in Sec. \ref{sec:footholdPlanning} with a \gls{CoM} motion planning that aims at always keeping the \gls{CoM} projection inside the scaled friction region $\mathcal{\hat{Y}}_f$;
\item \textit{feasible-region based walk:} this approach uses the same foothold strategy as above but makes sure that the \gls{CoM} projection always lies inside the scaled feasible region $\mathcal{\hat{Y}}_{fa}$ rather than $\mathcal{\hat{Y}}_f$. In this way therefore both friction and actuation constraints are explicitly considered at the motion planning level and are continuously re-planned for with a receding horizon of one step. 
\end{enumerate}

Evaluating the performance of these two strategies using the feasibility margin $r$ would skew the results in favor of the latter method, considering that the planner always makes sure that there exists a minimum feasible margin $r$ itself. For the assessment of the two planners' performance we therefore define the minimum joint-torque margin $m_{\tau}$. This corresponds to the minimum distance between the torque of each joint of the robot and their corresponding maximum and minimum values:
\begin{equation}
m_{\tau} = \min(d_0, \dots, d_n) 
\end{equation}
where:
\begin{equation}
d_i = min(\tau^{max}_i - \tau_i, \tau_i - \tau^{min}_i), \quad i = 0, \dots n
\end{equation}
The quantity $m_{\tau}$ measures how well the proposed online motion planner is able to keep the joint-torques away from their limits, while navigating complex geometry environments, being able to reach to the user direction commands or to unexpected disturbances.

It is important to mention that we evaluate $m_{\tau}$ only during during the triple support phases (\ie when only three legs are in contact with the ground and the fourth leg is in swing). This is because the triple support phase is the most critical for joint-torque limits (all the robot's weight is loaded on three legs rather than four) and because, as a consequence, the \gls{CoM} planning strategy optimizes the position of the \gls{CoM} only for this phase. Because of the static assumption that we assumed in ~\eqref{eq:staticForcepolytope}, the feasible region computation is only valid when the velocity of the robot's base is zero, condition which is not respected during the four-stance phase (\ie when the robot's base moves).

The values of $m_{\tau}$ for the two simulations are reported in the upper plot of Fig. \ref{fig:minJointTorque}. The red line shows the evolution of $m_{\tau}$ in the case of the friction region-based planning over the entire simulation (up to $14s$). The blue line shows instead the evolution of $m_{\tau}$ in the case of the feasible region-based planning over the entire simulation (up to $21s$). The recording of both simulations is stopped when the robot steps down the pallet with all four legs, the different duration of the simulations is therefore due to the different behavior they present during the negotiation of the pallet. We can notice that the minimum joint torque margin reached by the friction region based simulation of \SI[inter-unit-product =\ensuremath{\cdot}]{35}{\newton\meter} (dashed red line) occurs towards the conclusion of the experiment when the robot steps down the pallet with the last leg. The feasible region based walk instead performs an increased number of shorter steps before stepping down the pallet, in this way the simulation lasts longer and the minimum joint-torque margin of \SI[inter-unit-product =\ensuremath{\cdot}]{39}{\newton\meter} (dashed blue line) is higher than the simulation where only friction was considered.

\noindent%
\begin{minipage}{\linewidth}% to keep image and caption on one page
	\makebox[\linewidth]{%        to center the image
		\includegraphics[keepaspectratio=true,scale=0.075]{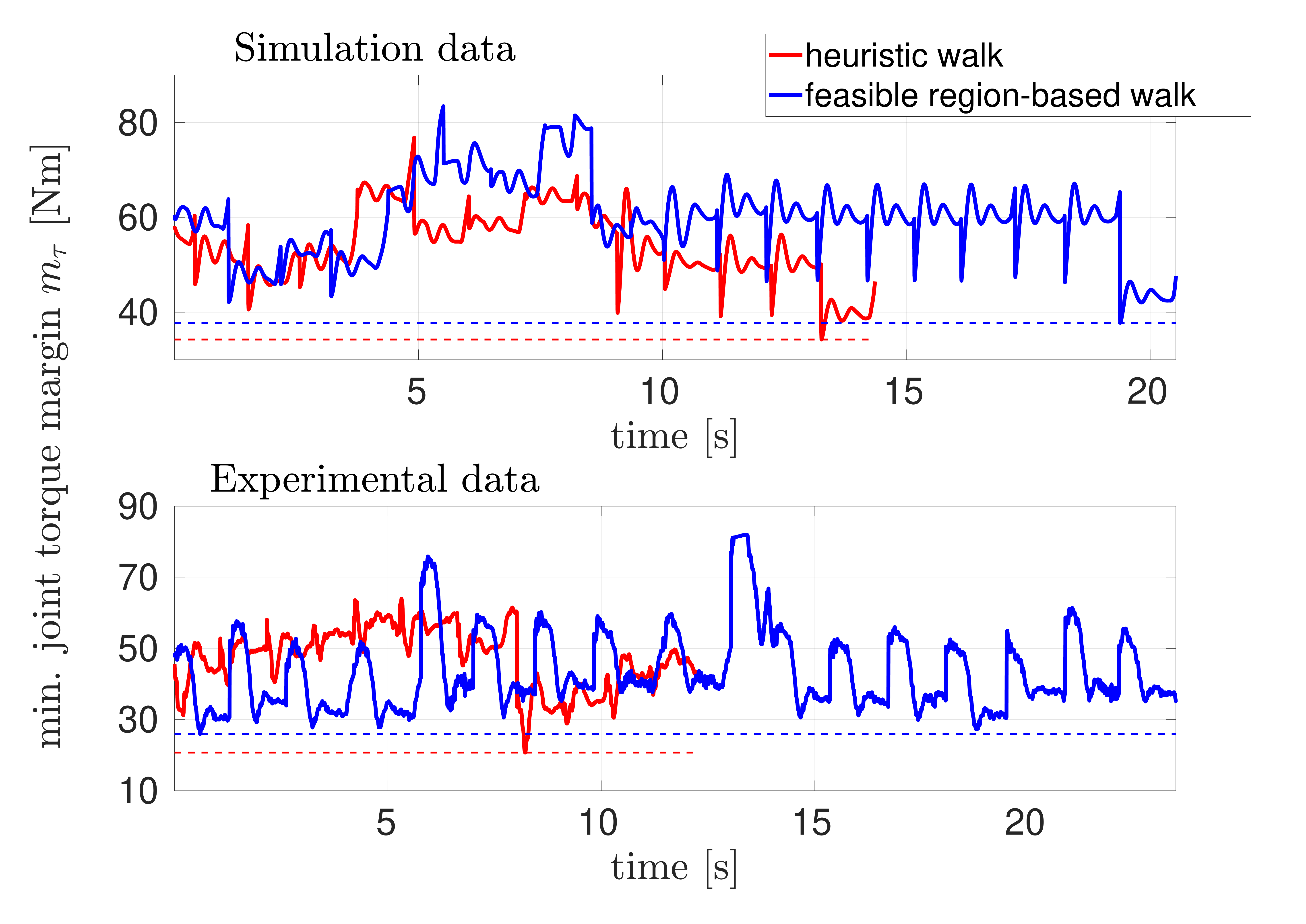}}
	\captionof{figure}[]{Minimum joint-torque margin $m_{\tau}$ in simulation (above) and hardware experiments (below). The red line shows the values of $m_{\tau}$ for a static crawl based on a heuristic strategy (red) and based on our proposed feasible region (blue). We can see that in both plots the minimum value of $m_{\tau}$ is higher when the strategy based on the feasible region is employed. This also results in shorter steps of the robot and thus a longer experiment time.}
	\label{fig:minJointTorque}
	\vspace{0.2cm}
\end{minipage}
The lower plot of Fig. \ref{fig:minJointTorque} refers instead to a hardware experiment where the HyQ robot walks over a moderately rough terrain made of bricks and plastic tiles while also carrying a \SI[inter-unit-product =\ensuremath{\cdot}]{10}{\kilogram} extra load on its trunk ({snapshots of this test can be seen in Fig.~\ref{fig:timelapse}}). {As the robot proceeds with a predefined foothold sequence and timing, the sample-based motion planner allows the robot to adapt its footholds and the trajectory of the \gls{CoM} to the height map received by the perceptive module. Having a limited number of sampled footholds $p$ allows one to set an upper bound on the maximum computation time required by the planner. In the considered experiment we had $p = 9$ which results in 9 sequential repetitions of the \gls{IP} algorithm for an overall computation time of about $63ms$ (every evaluation of a triple-stance phase takes in average $7ms$). This enables the replanning at a frequency of about \SI[inter-unit-product =\ensuremath{\cdot}]{15}{\hertz} in order to always include the latest values of the height map or possible changes in the state of the robot\footnote{{These computational performance would be significantly further improved by a parallelization of each foothold evaluation.}}.}

Also in this case, as in the simulation, the feasible region based approach presents a higher minimum joint-torque margin of \SI[inter-unit-product =\ensuremath{\cdot}]{29}{\newton\meter} (dashed blue line) compared to the \SI[inter-unit-product =\ensuremath{\cdot}]{21}{\newton\meter} margin that we measured for the friction region based approach (dashed red line). 
The video of the hardware experiments can be found in the accompanying video.
\section{Discussion}\label{sec:discussion}
In this Section we briefly address a few important aspects of the proposed feasible region such as its possible usage for dynamic gaits, whole-body control performance evaluation and for applications with bilateral contacts:
\begin{itemize}
		\item \textbf{Relaxation of the quasi-static assumption:} a \textit{dynamic} extension of the feasible region, that considers current configuration, velocity and acceleration of the robot, could be obtained performing the two following actions:
		\begin{enumerate}
			\item solve Alg.~\ref{alg:iterativeProjectionWithActuation} employing the \emph{dynamic} wrench polytopes, given in (\ref{eq:dynamicForcepolytope}), rather than the \emph{static} wrench polytopes given in (\ref{eq:staticForcepolytope});
			\item include the fictitious forces acting on the robot's \gls{CoM} inside the vector $\vc{u}$ in (\ref{eq:IPmatrices})\footnote{{This corresponds to \textit{slicing} the \gls{FWP} with a plane orthogonal to the aggregated centroidal wrench (\ie gravito-inertial wrench) \cite{Orsolino2018}}}.
		\end{enumerate}
{The above can be carried out given the current configuration, velocity and acceleration of the robot for visualization purposes or for testing the feasibility of the current state. Please note that this would also allow to explicitly consider the dependence of electric actuators on the joint velocity.
However, using this dynamic extension of the feasible region for \text{online} motion planning purposes is more challenging because of the curse of dimensionality. One possibility for achieving semi-dynamic motions would consists in employing the \emph{static} feasible region, scaled by a factor $s$ to enforce some robustness margin \cite{Audren2017}, in combination with a \emph{dynamic} reference point such as the \gls{zmp} or the \gls{icp};
\item \textbf{Benchmarking of whole-body controllers:} Having the \gls{CoM} inside the feasible region $\mathcal{Y}_{fa}$ provides a formal guarantee of existence of, at least, one set of contact forces that, simultaneously, satisfies the wrench polytope constraints and the friction cone constraints. $\mathcal{Y}_{fa}$ can thus also play a role in the evaluation of the performance of whole-body controllers. The interest is of this approach lies in the fact that constraints with different units ([\SI[inter-unit-product =\ensuremath{\cdot}]{}{\newton}] for friction force and [\SI[inter-unit-product =\ensuremath{\cdot}]{}{\newton\meter}] for joint-torques) are naturally mapped into a single number (in [\SI[inter-unit-product =\ensuremath{\cdot}]{}{\meter}]);
\item \textbf{Bilateral contacts:} actuation-consistent regions $\mathcal{Y}_{a}$ can be useful in other fields of robotics, such as loco-manipulation tasks, or whenever in presence of bilateral contacts (\eg climbing robots with magnetic grippers \cite{Eich2011}), heavy-duty walking machines with footsteps anchored to predefined locations (\eg the TITAN series robots \cite{Hodoshima2004}) and whenever a gripper/hand attached to the base link might lift an external weight and thus change the robot's loading conditions.
		}
	\end{itemize}

\section{Conclusion}\label{sec:conclusion}
In this paper we presented an approach for projecting joint-torque constraints from the high dimensional 
joint space ($n + 6$ \gls{DoFs}) of legged robots to the two-dimensional subspace of the \gls{CoM} {horizontal} plane. Despite the static assumption and the local nature of the resulting 2D {\textit{feasible region}, this strategy introduces the possibility to embed torque limitations of the actuators inside the formulation of motion planners based on simplified dynamic models and thus to efficiently plan realistic locomotion at high frequencies.}

{The feasible region $\mathcal{Y}_{fa}$ does not suffer from limitations related to specific robot morphologies or specific terrains (\eg flat terrains). As a consequence, this friction-and-actuation consistent area $\mathcal{Y}_{fa}$ can be employed for motion planning of legged robots on rough and complex terrains (the only limiting assumption being of static locomotion), where classical simplified models fail.}

Thanks to the computational efficiency of the feasible region $\mathcal{Y}_{fa}$ estimation, actuation-consistency and \textit{robustness} can be tested online at a minimum of \SI[inter-unit-product =\ensuremath{\cdot}]{100}{\hertz} rate and without any approximation regarding the location and orientation of the contacts. This last point allows our approach to be embedded in a map-based foothold optimization strategy that samples at \SI[inter-unit-product =\ensuremath{\cdot}]{15}{\hertz} feasible footholds on the height map provided by the vision module (see Fig. \ref{fig:footholdPlanning}). 

We reported simulations of the \gls{HyQ} robot crossing a pallet of \SI[inter-unit-product =\ensuremath{\cdot}]{0.15}{\meter} and hardware experiments of the real robot on a terrain of mild roughness (with bricks up to \SI[inter-unit-product =\ensuremath{\cdot}]{0.1}{\meter}) while carrying an additional load of \SI[inter-unit-product =\ensuremath{\cdot}]{10}{\kilogram}. The foothold strategy corrected the steps (\eg making them smaller) when needed to increase the robustness to uncertainty and to disturbances represented by the feasibility margin $r$. 

{In the two following we present a few possible future developments of the work presented in this paper:
\begin{itemize}
	\item \textbf{Global feasible region:} as we discussed in this paper the feasible region is a configuration dependent quantity, whose validity is thus guaranteed only in a \textit{local} neighborhood of the considered state. Our future efforts will focus on the computation of a \textit{global} extension of the feasible region which only depends on the contacts' pose;
	%\item \textbf{Closed-form computation:} analytically determining the feasible region for a given set of contacts represents a tantalising alternative to the \gls{IP} algorithm. This could indeed possibly involve a further computational speed up and it could also enhance the analytic solution of the inverse problem, \ie find the optimal contact point for a predefined feasible region;
	\item \textbf{Offline pre-computation and differentiation:} it has been shown in the past that \gls{CoM}-admissible regions that respect feasibility constraints such as kinematic limits and collisions can be learned using proper function approximators \cite{carpentier2017}. A similar approach could be employed to learn an approximate \textit{differentiable} relationship between the robot's state and the feasible region. This would allow us to embed the feasible region as a constraint within, for example, a gradient descent trajectory optimization problem.
\end{itemize}

\section{Appendix}
We recall in this Appendix few of the main concepts and definitions connected to computational geometry that are heavily used in this manuscript. Most definitions are taken from the following sources \cite{Fukuda1998, Delos2014, Delos2015, Borrelli2017}.

\subsection{Generic Bounded and Unbounded Polyhedra Definitions}\label{appendix:boundedPolygonsAndCones}
	 
Main definitions and terminology used in sets representation and adopted in this paper:
	\begin{itemize}
		\item A convex \textit{polyhedron} $\mathcal{H}$ is a subset of $\Rnum^d$ that solves a finite set of $m$ linear inequalities. The volume of a polyhedron can therefore be either bounded or unbounded. This is a generic definition that may include both (bounded) polytopes and (unbounded) polyhedral cones.
		\begin{equation}
		\mathcal{H} = \{ \vc{x} \in \Rnum^d \quad | \quad \vc{A}\vc{x} \leq \vc{b} \}
		\end{equation}
		with $\vc{A} \in \Rnum^{m \times d}$ and $\vc{b} \in \Rnum^{m}$.
		
		\item A convex \textit{polytope} $\mathcal{P}$ is a subset of $\Rnum^d$ that solves a finite set of $m$ linear inequalities and is bounded.
		\begin{equation}
		\mathcal{P} = \{ \vc{x} \in \Rnum^d \quad | \quad \vc{A}\vc{x} \leq \vc{b} \}
		\end{equation}
		with $\vc{A} \in \Rnum^{m \times d}$ and $\vc{b} \in \Rnum^{m}$.
		
		\item A convex \textit{polygon} $\mathcal{P}$ is a polytope in dimension $d = 2$:
		\begin{equation}
				\mathcal{P} = \{ \vc{x} \in \Rnum^2 \quad | \quad \vc{A}\vc{x} \leq \vc{b} \}
				\end{equation}
				with $\vc{A} \in \Rnum^{m \times 2}$ and $\vc{b} \in \Rnum^{m}$.
		
		\item A convex \textit{zonotope} $\mathcal{Z}$ is a special kind of polytope in $\Rnum^d$ that presents particular symmetry with respect to the its center \cite{Althoff2010, Althoff2016}. A zonotope can therefore be fully described by its center $\vc{c} \in \Rnum^d$ and its $p$ generators $\vc{g} \in \Rnum^d$.
		\begin{equation}
		\mathcal{Z} = \Big\{ \vc{c} + \sum_{i=1}^{p} \alpha_i \vc{g}_i \quad | \alpha_i \in [-1, 1],  \vc{g}_i \in \Rnum^d, \vc{c} \in \Rnum^d \Big\}
		\end{equation}
		
		\item A convex \textit{polyhedral cone} $\mathcal{C}$ is a subset of $\Rnum^d$ that solves a finite set of $m$ linear inequalities. Geometrically, each linear inequality defines a hyperplane that has to pass through the origin.
		\begin{equation}
		\mathcal{C} = \{ \vc{x} \in \Rnum^d \quad | \quad \vc{C}\vc{x} \leq \vc{0} \}
		\label{eq:coneHrep}
		\end{equation}
		with $\vc{A} \in \Rnum^{m \times n}$ and $\vc{0} \in \Rnum^{m}$ is a null vector.
	\end{itemize} 

Convex polyhedra, polytopes, zonotopes and cones are called $d$-polyhedra ($d$-polytopes, $d$-zonotopes or $d$-cones) if they have a non-zero interior in $\Rnum^d$; 

%TODO check this  they cannot have more that d dimensions
%$d$-polyhedra will have a null interior in all dimensions larger than $d$ (in which case they are nicknamed \textit{flat} polyhedra/polytopes/cones).\\
%In this paper we will drop the adjective convex referred to these polyhedra for compactness. All the mentioned polyhedra are therefore to be considered to be convex unless explicitly specified.

In the computational geometry terminology, a hyperplane $h$ of $\Rnum^d$ is a supporting hyperplane of the polyhedron $\mathcal{H}$ if one of the closed halfspaces of $h$ contains $\mathcal{H}$. A \textit{face} $\mathcal{F}$ of $\mathcal{H}$ is a generic term to indicate either an empty set, $\mathcal{H}$ itself or the intersection between $\mathcal{H}$ and a supporting hyperplane. The \textit{faces} of dimension $0, 1, d-1$ and $d-2$ are usually named \textit{vertices, edges, ridges} or \textit{facets} \cite{Fukuda1998}.

\begin{itemize}
\item A \textit{half-space} is either of the two parts in which a hyperplane divides an affine space.
\item A \textit{generator} is a broad term to indicate all the elements of Euclidean space $\Rnum^d$ that can be used to represent the considered set. Depending on the considered type of polyhedron, generators may include \textit{vertices}, \textit{rays} (or \textit{edges}).% and \textit{intervals}.
\end{itemize}
According to the Minkowski-Weil theorem \cite{Delos2015}, polyhedra can be equivalently described in terms of their {half-spaces ($\mathcal{H}$-description) or in terms of their generators ($\mathcal{G}$-description). The generators, depending on the considered geometrical object, can consist of vertices ($\mathcal{V}$-description), rays ($\mathcal{R}$-description) or an interval ($\mathcal{I}$-description).} Polytopes, for example, can be equivalently described in terms of $\mathcal{H}$- and/or $\mathcal{V}$-description. Polyhedral cones $\mathcal{C}$ can be equivalently described in terms of $\mathcal{H}$-description (see ~\eqref{eq:coneHrep}) and/or $\mathcal{R}$-description:
\begin{equation}
\mathcal{C} = \Big\{ \sum_{i=1}^{p} \alpha_i \vc{r}_{i} \quad | \quad \forall \alpha_i \geq 0, \quad  \sum_{i=1}^{p} \alpha_i = 1, \quad  \vc{r}_{i} \in \mathcal{R} \Big\}
\end{equation}
where $p$ is the number of rays of the set of rays $\mathcal{R}$:
\begin{equation}
\mathcal{R} = \left\{ \vc{r}_{1}, \dots, \vc{r}_{p} \quad | \quad \vc{r}_{i} \in \Rnum^d \right\}
\end{equation} A cone, however, can not be represented by $\mathcal{V}$-description as it only owns one vertex which is placed in the origin of the reference frame. 

\subsection{Minkowsky Sums and Convex Cones}\label{appendix:minkowskiSums}
In the following we will discuss the main properties of sum of sets and convex hull algorithm:
\begin{itemize}
	\item Given two convex sets $\mathcal{A}$ and $\mathcal{B}$, their addition (called \textit{Minkowski sum}), indicated by the operator $\oplus$, another set is defined as the sum of the all elements of $\mathcal{A}$ with all the elements of $\mathcal{B}$:
	\begin{equation}
	\mathcal{A} \oplus \mathcal{B} = \{ \vc{a} + \vc{b} \quad | \quad \vc{a} \in \mathcal{A}, \quad \vc{b} \in \mathcal{B} \}
	\end{equation}
	which presents a $\mathcal{O}(a \cdot b)$ time (where $a$ is the cardinality of $\mathcal{A}$ and $b$ is the cardinality of $\mathcal{B}$).
	
	\item For a given convex set $\mathcal{S} = \{ \vc{s}_1,  \dots, \vc{s}_n | \vc{s} \in \Rnum^d \}$ composed of $n$ finite elements of dimension $d$, their \textit{convex hull} is defined as the set of all the convex combinations of all its elements (e.g. the vertices of a polytope):
	\begin{equation}
	ConvHull(S) = \Big\{ \sum_{i=1}^{n} \alpha_i \vc{s}_i \quad | \quad \forall \alpha_i \geq 0, \quad  \sum_{i=1}^{n} \alpha_i = 1 \Big\}
	\end{equation}
	
\end{itemize}

The convex hull distributes over the Minkowski sum, meaning that the following property holds:
\begin{equation}
	ConvHull(\mathcal{A} \oplus \mathcal{B}) = ConvHull(\mathcal{A}) \oplus ConvHull(\mathcal{B})
\end{equation}
In the worst-case output the complexity of the problem is $\mathcal{O}(n^{[d/2]})$.
% although there exist sophisticated algorithms that can compute the convex hull in time $\mathcal{O}(n \log(h))$ where $h$ is the number of points of the resulting convex hull.

For the computation of many locomotion related geometrical objects, such as the \gls{cwc}, it is important to notice that, given the $\mathcal{R}$-representation of two polyhedral cones $\mathcal{C}_1$ and $\mathcal{C}_2$:
\begin{equation}
 \begin{aligned}
\mathcal{C}_1 =\Big\{ \sum_{i=1}^{p_1} \alpha_i \vc{r}_{1,i} \quad | \quad \forall \alpha_i \geq 0, \quad  \sum_{i=1}^{p_1} \alpha_i = 1, \quad  \vc{r}_{1,i} \in \mathcal{R}_1 \Big\}
\\
\mathcal{C}_2 = \Big\{ \sum_{i=1}^{p_2} \alpha_i \vc{r}_{2,i} \quad | \quad \forall \alpha_i \geq 0, \quad  \sum_{i=1}^{p_2} \alpha_i = 1, \quad  \vc{r}_{2,i} \in \mathcal{R}_2 \Big\}
 \end{aligned}	
\end{equation}
the $\mathcal{R}$-representation of their Minkowski sum $\mathcal{C}_{sum}$ can be obtained by stacking together (i.e., using the \textit{union} operator $\cup$) the set of rays $\mathcal{R}_1$ and $\mathcal{R}_2$ of the two individual cones:
\begin{equation}
\begin{aligned}
\mathcal{C}_{sum} = & \quad  \mathcal{C}_1 \oplus \mathcal{C}_2 = \Big\{ \sum_{i=1}^{p_1 + p_2} \alpha_i \vc{r}_{i} \quad | \quad \forall \alpha_i \geq 0, \\
& \quad \sum_{i=1}^{p_1 + p_2} \alpha_i = 1, \quad  \vc{r}_{i} \in \mathcal{R}_1 \cup \mathcal{R}_2 \Big\}
\end{aligned}
\end{equation}
Despite yielding a redundant representation with internal rays, this property allows a considerable speed-up ($\mathcal{O}(p_1+p_2)$ time) compared to the Minkowski sum of two convex bounded polytopes ($\mathcal{O}(p_1 \cdot p_2)$ time).


%\bibliographystyle{IEEEtran}
%\bibliography{refs/references.bib}

\begin{IEEEbiography}[{\includegraphics[width=1in,height=1.25in,clip,keepaspectratio]{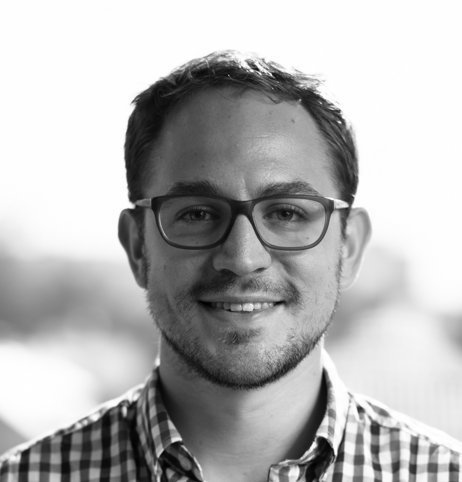}}]{Romeo Orsolino}
completed his Bsc. in Mech. Eng. in 2013 and his Msc. in Robotics Eng. in 2015. In the same year, he then joined the DLS team at IIT for a PhD focusing on motion planning for legged locomotion in rough terrains. After successfully defending his PhD thesis in February 2019, Romeo remained at the DLS lab as a postdoc until September 2019. Since October 2019, he is now a post doctoral researcher in the Dynamic Robot Systems (DRS) team of the Oxford Robotics Institute (ORI) where he continues his work on multi-contact motion planning, optimal control, dynamics and perception.
\end{IEEEbiography}

\begin{IEEEbiography}[{\includegraphics[width=1in,height=1.25in,clip,keepaspectratio]{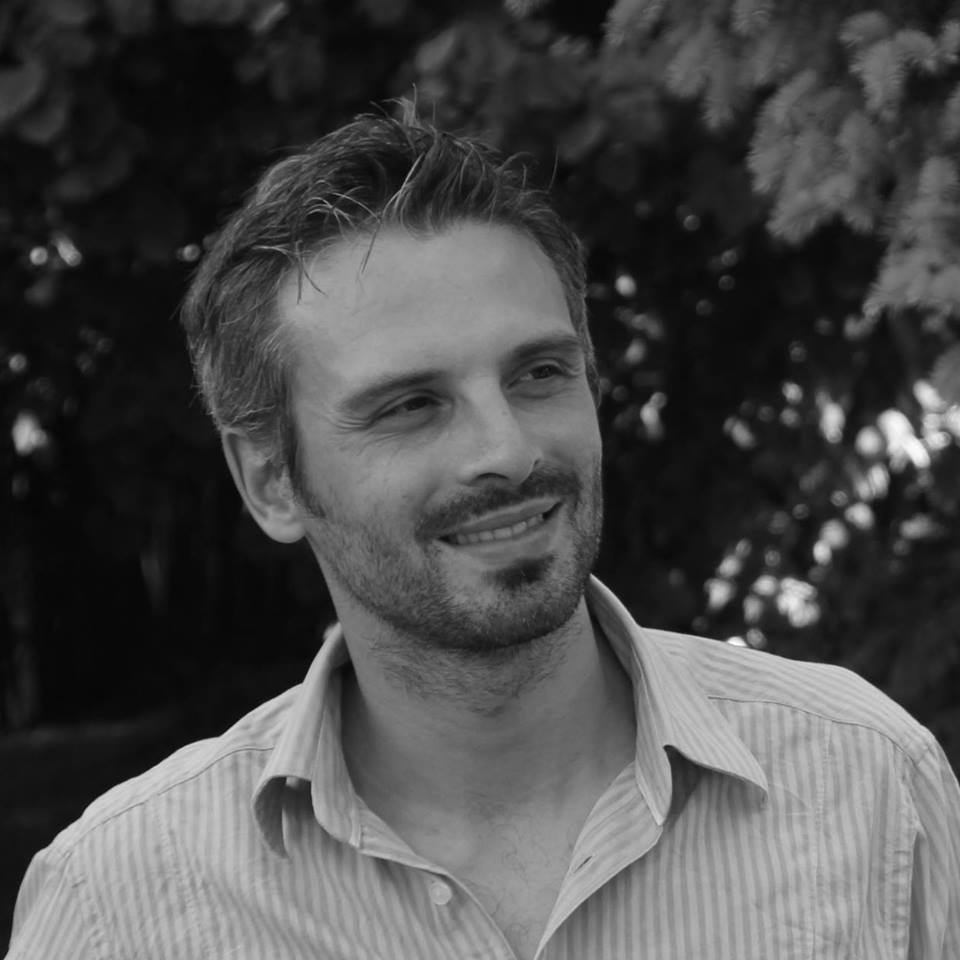}}]{Michele Focchi}
is currently a Researcher at the DLS team in IIT. He received both his Bsc. and the Msc. in Control System Engineering from Politecnico di Milano. After gaining some R$\&$D experience in the industry, in 2009 he joined IIT where he developed a micro-turbine for which he obtained an international patent. In 2013, he got a PhD in robotics, getting involved in the Hydraulically-actuated Quadruped (HyQ) robot project.  Currently his research interests focus on pushing the performance of quadruped robots in traversing unstructured environments, by using optimization-based planning strategies to perform dynamic planning.
 
\end{IEEEbiography}

\begin{IEEEbiography}[{\includegraphics[width=1in,height=1.25in,clip,keepaspectratio]{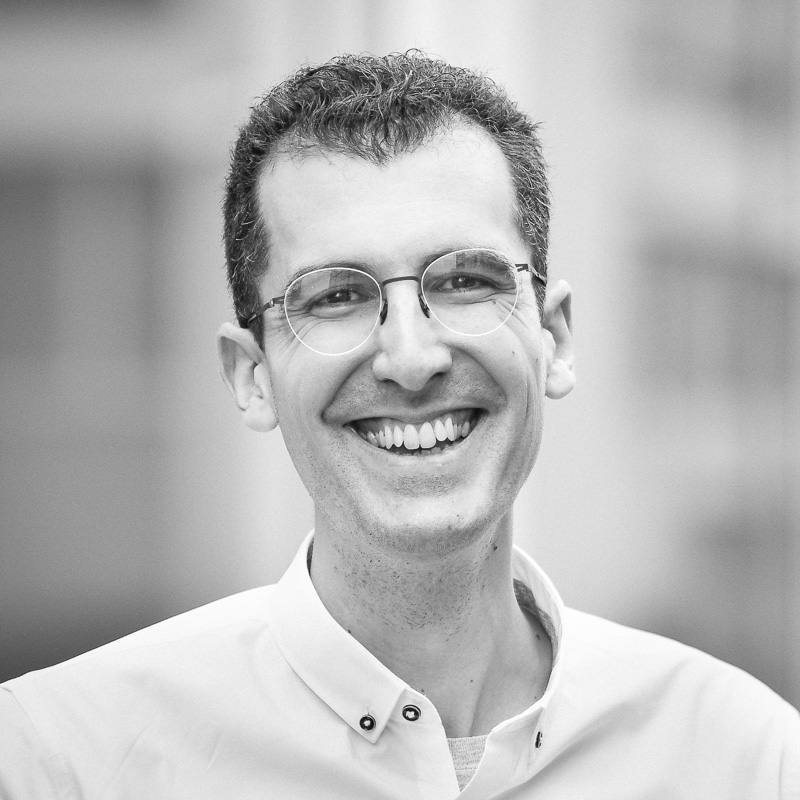}}]{St{\'e}phane, Caron} is a Locomotion Software Engineer at ANYbotics. He received his MSc in 2012 from the \'{E}cole Normale Sup\'{e}rieure of Paris and his PhD in 2016 from the University of Tokyo with a dissertation on multi-contact stability conditions. After joining CNRS in 2017, he studied balance control with height variations and developed an open-source walking and stair climbing controller for the HRP-4 humanoid. St\'{e}phane also takes projects to the field, such as valve turning (2015 DARPA Robotics Challenge) or walking up stairs at the Airbus Saint-Nazaire factory (2019 COMANOID Project). He is interested in legged robot locomotion.
	
\end{IEEEbiography}

\begin{IEEEbiography}[{\includegraphics[width=1in,height=1.25in,clip,keepaspectratio]{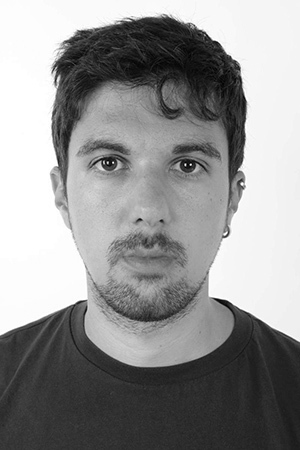}}]{Gennaro Raiola}
received his Bsc. and Msc. degrees (magna cumlaude) in computer and automation engineering at the University of Naples “Federico II” in 2010 and 2012 respectively. He obtained his PhD degree in robotics from the University of Paris-Saclay, France, in February 2017. He worked as research engineer at PAL Robotics and as post doctoral Researcher at the Robotics and Mechatronics group, University of Twente, The Netherlands. After spending one year as a Postdoctoral Researcher at the DLS team in IIT (Italy) he then moved to Nasa where he is currently working on robotics for space assembly tasks.
\end{IEEEbiography}

\begin{IEEEbiography}[{\includegraphics[width=1in,height=1.25in,clip,keepaspectratio]{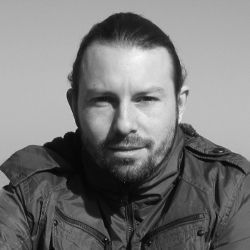}}]{Victor Barasuol}
was born in São Miguel do Oeste/SC, Brazil. He holds 
a Diploma in Electrical Engineering from Universidade do Estado 
de Santa Catarina - UDESC (2006). He has master degree in 
Electrical Engineering (2008) and a doctorate degree in Automation 
and Systems Engineering (2013) from Universidade Federal de 
Santa Catarina - UFSC. Victor Barasuol is currently a senior postdoctoral  
researcher of the Dynamic Legged Systems lab (DLS) at the Istituto 
Italiano di Tecnologia (IIT). His main research topics are motion 
generation and control for quadruped robots with emphases 
in dynamic locomotion and reactive actions using proprioceptive
and exteroceptive sensors. 
\end{IEEEbiography}

\begin{IEEEbiography}[{\includegraphics[width=1in,height=1.25in,clip,keepaspectratio]{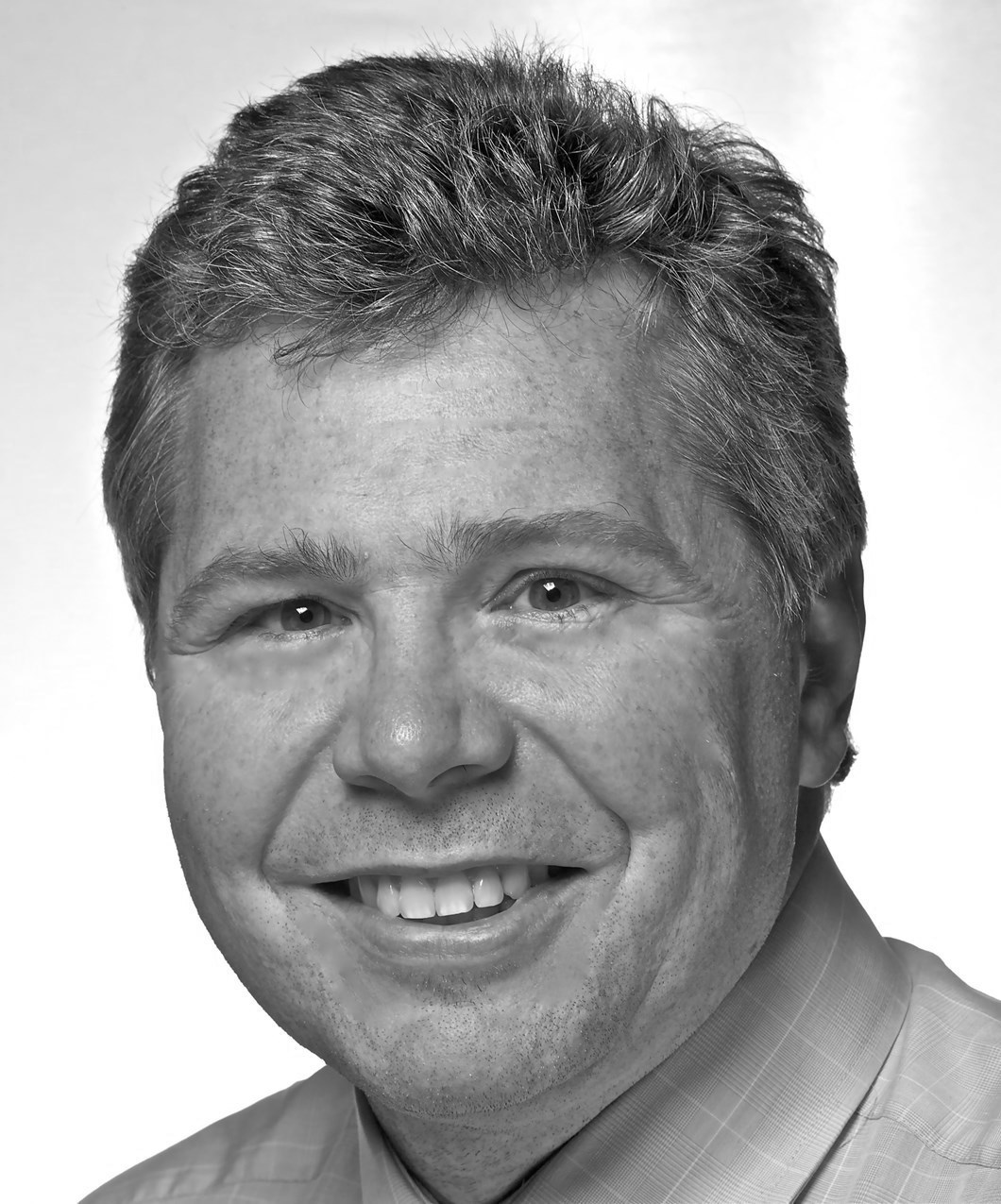}}]{Darwin G. Caldwell}
received his BSc. and PhD in Robotics from the University of Hull in 1986 and 1990 respectively. In 1994 he received an MSc in Management from the University of Salford. From 1989 to 2006 he was at the University of Salford as a Lecturer, S.Lecturer, and Reader, becoming Full Professor of Advanced Robotics in 1999. He is an Honorary Professor at the Universities of Manchester, Sheffield, Bangor, and Kings College London in the UK and Tianjin University and Shenzhen Academy of Aerospace Technology in China. In 2015 he was elected a Fellow of the Royal Academy of Engineering.
\end{IEEEbiography}

\begin{IEEEbiography}[{\includegraphics[width=1in,height=1.25in,clip,keepaspectratio]{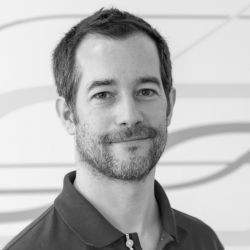}}]{Claudio Semini}
received an Msc. degree in electrical engineering and information technology from
ETH Zurich, Switzerland, in 2005. He is currently
the Head of the Dynamic Legged Systems (DLS)
Laboratory at Istituto Italiano di Tecnologia (IIT).
From 2004 to 2006, he first visited the Hirose
Laboratory at Tokyo Tech, and later the Toshiba
R$\&$D Center, Japan. During his doctorate at the IIT, he developed the hydraulic
quadruped robot HyQ. After a postdoc with the same department, in 2012, he
became the head of the DLS lab. His research interests include the construction
and control of versatile, hydraulic legged robots for real-world environments. 
\end{IEEEbiography}

\end{document}